%% file: neurips_2026.tex
\documentclass{article}

\PassOptionsToPackage{numbers, compress}{natbib}

\usepackage[main, final]{neurips_2026}

\usepackage{amsthm}

\input{math_commands}

\input{shortcut}

\usepackage[utf8]{inputenc} %
\usepackage[T1]{fontenc}    %
\usepackage{hyperref}       %
\usepackage{url}            %
\usepackage{booktabs}       %
\usepackage{amsfonts}       %
\usepackage{nicefrac}       %
\usepackage{microtype}      %
\usepackage{xcolor}         %

\title{Contrastive Distribution Matching for Amortized Sequential Monte Carlo in Discrete Diffusion}

\author{%
  \stepcounter{footnote}%
  \textbf{Jaihoon Kim}\textsuperscript{\textnormal{1}}\thanks{Correspondence to: \texttt{jh27kim@kaist.ac.kr}} \hspace{2em}
  \textbf{Taehoon Yoon}\textsuperscript{\textnormal{2}} \hspace{2em}
  \textbf{Prin Phunyaphibarn}\textsuperscript{\textnormal{1}} \\[1.75ex]
  \textbf{Seungjun Kim}\textsuperscript{\textnormal{1}} \hspace{2em}
  \textbf{Morteza Mardani}\textsuperscript{\textnormal{3}} \hspace{2em}
  \textbf{Minhyuk Sung}\textsuperscript{\textnormal{1}} \\[1.75ex]
  \textsuperscript{\textnormal{1}}KAIST \qquad \textsuperscript{\textnormal{2}}University of Michigan \qquad \textsuperscript{\textnormal{3}}NVIDIA \\[1.5ex]
  Project Page: \texttt{https://cdm-smc.github.io/} %
}

\newif\ifpaper
\paperfalse

\begin{document}

\maketitle

\input{Sections/00_Abstract}

\input{Sections/01_Introduction}

\input{Sections/02_Background}
\input{Sections/03_Method}

\input{Sections/04_Related_Works}

\input{Sections/05_Experiments.tex}
\input{Sections/06_Conclusion.tex}

\clearpage
\newpage

\section*{Acknowledgments}
We thank Jason Yoo for insightful discussions on learning the value function for SMC in generative models, and Yuchen Zhu for providing new insights into the role of negative gradients. 
This work was supported by the NRF grant (RS-2026-25486000); IITP grants (RS-2024-00399817, RS-2025-25441313, RS-2025-25443318, RS-2026-25526850, RS-2026-25617720) funded by MSIT, Korea; the Industrial Technology Innovation Program grant (RS-2025-02317326) funded by MOTIE, Korea; the InnoCORE program of MSIT (AI Meta-Scientist, N10260110); and the DRB-KAIST SketchTheFuture Research Center.

{\small
\bibliographystyle{plainnat}
\bibliography{main}
}

\ifpaper
\else
    \clearpage
    \newpage
    \onecolumn
    \appendix
    \setcounter{section}{0}
    \def\thesection{\Alph{section}}
    \section*{Appendix}
    \input{Sections/A_Appendix}

\fi

\clearpage
\newpage
\input{checklist}

\end{document}

%% file: math_commands.tex
\usepackage{amsmath,amsfonts,bm}

\def\eqref#1{equation~\ref{#1}}

\def\1{\bm{1}}

\DeclareMathAlphabet{\mathsfit}{\encodingdefault}{\sfdefault}{m}{sl}
\SetMathAlphabet{\mathsfit}{bold}{\encodingdefault}{\sfdefault}{bx}{n}

%% file: shortcut.tex
\usepackage[linesnumbered,ruled,vlined]{algorithm2e}
\usepackage{setspace}
\usepackage{amsfonts}       %
\usepackage{booktabs}       %
\usepackage{enumitem}
\usepackage[T1]{fontenc}    %
\usepackage[utf8]{inputenc} %
\usepackage{makecell}
\usepackage{microtype}      %
\usepackage{multicol,multirow}
\usepackage{nicefrac}       %
\usepackage{tabularx}
\usepackage{url}            %

\usepackage{longtable}
\usepackage{filecontents,catchfile,pgffor}

\usepackage{environ}
\usepackage[percent]{overpic}
\usepackage[dvipsnames]{xcolor}

\usepackage{ragged2e}   %
\usepackage{float}      %

\newcolumntype{L}{>{\RaggedRight\arraybackslash}X}

\usepackage{amsmath}
\usepackage{amssymb}
\usepackage{booktabs}
\usepackage{dsfont}
\usepackage{graphicx}
\usepackage{makecell}
\usepackage{multirow}
\usepackage{pifont}%
\usepackage[group-separator={,}]{siunitx}
\usepackage{tabularx}
\usepackage{caption}
\usepackage{soul}
\usepackage{changepage,threeparttable}
\usepackage{colortbl}
\usepackage[accsupp]{axessibility}  %
\usepackage{mathtools}
\usepackage{xfrac}
\usepackage{enumitem}
\usepackage{catchfile}
\usepackage{wrapfig}

\usepackage{longtable}
\usepackage{tabu}
\usepackage{tcolorbox}
\usepackage{supertabular}
\usepackage{caption}
\usepackage{subcaption}
\usepackage{pifont}
\usepackage{adjustbox}
\usepackage{titletoc}
\usepackage{multicol}
\newcolumntype{Y}{>{\centering\arraybackslash}X}
\usepackage{physics}

\usepackage{hyperref}       %
\usepackage[capitalize,nameinlink]{cleveref}

\usepackage{CJKutf8}

\usepackage{amsthm}

\newcommand{\B}{\mathbf}

\crefname{equation}{Eq.}{Eqs.}
\Crefname{equation}{Eq.}{Eqs.}
\crefname{figure}{Fig.}{Figs.}
\Crefname{figure}{Fig.}{Figs.}
\crefname{table}{Tab.}{Tabs.}
\Crefname{table}{Tab.}{Tabs.}
\crefname{section}{Sec.}{Secs.}
\Crefname{section}{Sec.}{Secs.}
\Crefname{section}{Sec.}{Secs.}
\crefname{algorithm}{Alg.}{Algs.}
\Crefname{algorithm}{Alg.}{Algs.}
\crefname{appendix}{Appendix.}{Appendix.}
\Crefname{appendix}{Appendix.}{Appendix.}

\newcommand{\Ours}[0]{\texttt{CDM}}
\newcommand{\Oursbf}[0]{\fontfamily{lmtt}\fontseries{b}\selectfont \Ours{}}

\newcolumntype{Y}{>{\centering\arraybackslash}X}

\newcommand{\svdd}[0]{Soft Value}

\newcommand{\ie}{i.e.,}
\newcommand{\eg}{e.g.,}

\definecolor{color_1}{RGB}{255,0,128}
\definecolor{color_2}{RGB}{128,128,0}
\definecolor{color_3}{RGB}{0,128,0}
\definecolor{color_4}{RGB}{128,0,0}
\definecolor{color_5}{RGB}{128,0,128}

\definecolor{Goldenrod}{RGB}{218, 165, 32}
\definecolor{ForestGreen}{RGB}{34, 139, 34}

\definecolor{highlight_color}{RGB}{255,0,128}
\definecolor{blue_highlight_color}{RGB}{0,128,255}
\definecolor{green_highlight_color}{RGB}{51,145,40}
\definecolor{orange_highlight_color}{RGB}{255,160,0}
\definecolor{purple_highlight_color}{RGB}{128,0,128}

\newcommand{\mycomment}[1]{}

%% file: Sections/00_Abstract.tex
\vspace{-1.0\baselineskip}
\begin{abstract}
\vspace{-0.5\baselineskip}
Discrete diffusion models have emerged as powerful frameworks for generating structured categorical data. 
However, efficiently sampling from reward-tilted distributions remains a fundamental challenge. 
While Twisted Sequential Monte Carlo (SMC) offers asymptotic exactness for this task, estimating the optimal twist function in discrete state spaces necessitates costly Monte Carlo approximations, resulting a severe computational bottleneck at inference. 
To overcome this limitation, we introduce Contrastive Distribution Matching (\Ours{}), a novel framework that amortizes the cost of SMC inference by learning a parameterized twist function via positive and negative samples. 
For efficient training, we reformulate the gradient estimator to leverage the closed-form forward kernels of discrete diffusion models. 
In practice, evaluating our learned twist function incurs less than $5\%$ additional computational overhead compared to a single forward pass of the base model. 
Through extensive empirical evaluations, we demonstrate that \Ours{} consistently outperforms existing baselines under matched wall-clock time. 
We validate the effectiveness and versatility of our approach across a diverse range of applications, including toxic text generation, regulatory DNA sequence design, protein designability, and diffusion large language model alignment. 
\end{abstract}

%% file: Sections/01_Introduction.tex
\section{Introduction}
\label{sec:introduction}

Diffusion models have demonstrated remarkable generative performance across a wide array of continuous domains~\cite{Rombach:2022LDM,flux2024}. 
Recently, their application to discrete state spaces has yielded significant breakthroughs; in language modeling, discrete diffusion models not only enable efficient few-step generation~\cite{potaptchik:2026DiscreteFlowMap, lee:2026FlowMapLM, park:2026pairflow} but also achieve sample quality on par with autoregressive language models~\cite{nie:2025llada,yang:2026mmada,ye:2025dream7b}. 
Furthermore, discrete diffusion has been successfully extended to scientific applications, driving advances in sequence design tasks such as regulatory DNA~\cite{Li2024:SVDD,sarkar:2024d3} and de novo protein~\cite{Wang:2025DPLM} generation. 

A central application of these models is \emph{reward alignment}. 
Given a scalar reward representing human preference~\cite{liu:2024skywork, wang:2024interpretable} or protein designability~\cite{Lin:2023ESM}, the objective is to sample from a tilted target distribution that biases the pretrained prior toward higher values of the downstream reward. 
To sample exactly from this target distribution, the optimal proposal is formulated by tilting the pretrained base model with an \emph{optimal twist function}. 

In this work, we integrate the Sequential Monte Carlo (SMC) framework~\cite{Doucet2001:SMC,Naesseth2019:SMC}, an asymptotically unbiased sampler, with discrete diffusion models. 
In the continuous domain, the SMC framework has been widely adopted for reward alignment largely due to its computational efficiency and empirical success~\cite{Wu:2023TDS, Cardoso2024:MCGDiff, Kim:2025DAS, singhal:2025general, skreta:2025feynman, ren:2025driftlite, albergo:2025nets}. 
This tractability stems from two key properties: one can easily construct locally optimal proposals using reward gradients~\cite{Chung:2023DPS, Kim:2025DAS}, and Tweedie's formula~\cite{efron:2011tweedie} provides a computationally efficient estimate of the clean state to approximate the twist function. 
In contrast, translating these successes to discrete domains presents new challenges. 
Since the state space is discrete, one needs to rely on Gumbel-Softmax trick~\cite{jang:2016categorical} to approximate locally optimal proposals, which often leads to gradient bias and optimization instability~\cite{paulus:2020gradient, liu:2023bridging}. 
More importantly, the absence of Tweedie's estimate in discrete diffusion~\cite{raphan:2011least} leaves Monte Carlo estimation as the standard practice for approximating the twist function~\cite{ou2025:smc_amot, Li2024:SVDD, dang:2025inference}, which can introduce a significant inference overhead when the downstream reward is computationally expensive (\eg~protein designability). 

Motivated by this bottleneck, we propose Contrastive Distribution Matching (\Ours{}), which learns the twist function via a contrastive learning objective to reduce the twist function evaluation to a \emph{constant-time} operation, amortizing SMC inference. 
In contrast to existing regression-based methods applied to discrete diffusion that learn the twist by drawing samples from a base proposal~\cite{Li2024:SVDD, Uehara:2024Finetuning}, \Ours{} minimizes the forward KL divergence against the target distribution. 
The gradient of this objective exhibits a contrastive structure, utilizing positive and negative samples to upweight high-reward regions while downweighting suboptimal ones. 
Additionally, we introduce a novel training scheme that leverages the forward process of the diffusion model for efficient training. Specifically, we maintain a buffer of positive samples drawn from the approximated target distribution and apply the closed-form forward kernel, allowing a single clean sample to be reused across multiple timesteps and gradient updates.

Our experimental evaluations demonstrate that \Ours{} consistently achieves superior scaling behavior, outperforming baselines in a diverse range of applications: toxic text generation, regulatory DNA design, protein generation, and diffusion LLM (dLLM) preference alignment. 
Furthermore, since \Ours{} learns the twist function, it is agnostic to the choice of the proposal distribution. 
This allows it to be paired with \emph{any} proposal distribution, including those already fine-tuned (\eg~d1~\cite{zhao2025:d1}, DRAKES~\cite{wang2024:drakes}), for further synergistic performance gains. 
Moreover, we demonstrate that the contrastive learning objective of \Ours{} yields superior performance and more efficient training compared to the standard regression-based twist objective~\cite{Uehara:2024Finetuning, Li2024:SVDD}. 

In summary, our key contributions are as follows:
\begin{itemize}[leftmargin=*, itemsep=2pt, topsep=2pt]
\item We propose Contrastive Distribution Matching (\Ours{}), an SMC-amortization framework for discrete diffusion that reduces the cost of applying the twist at inference time to a constant-time operation. 
\item We design a novel, diffusion-native training scheme that leverages the closed-form forward process, enabling efficient training that scales to expensive reward functions. 
\item We demonstrate the versatility of \Ours{} across a broad range of applications, including toxic text generation, regulatory DNA design, protein generation, and dLLM alignment, consistently showing  superior performance. 
\item We validate that \Ours{} delivers synergistic improvements even when paired with fine-tuning-based methods, while demonstrating superior efficacy than the regression-based twist objectives employed in previous discrete diffusion models~\cite{Li2024:SVDD}. 
\end{itemize}

%% file: Sections/02_Background.tex
{\vspace{-0.25\baselineskip}}
\section{Preliminary: Discrete Diffusion}
\label{sec:background}

Let $\mathcal{V} := \{\B{x} \in \{0,1\}^V : \sum_{i=1}^V \B{x}_i = 1\}$ define the space of $V$ category one-hot vectors. 
We write $\Delta^{V}$ for the $V$-simplex and $\text{Cat}(\cdot; \boldsymbol{\pi})$ for the categorical distribution with probability vector $\boldsymbol{\pi} \in \Delta^{V}$. 
A prominent class of generative models for discrete state spaces is the Masked Diffusion Model (MDM)~\cite{Sahoo:2024MDLM, campbell2022:continuous, shi2024:simplified}, which defines a forward corruption process terminating in a \emph{mask} state $\B{m}$.

Let $(p_t)_{t=0}^T$ denote the sequence of marginal distributions induced by this forward process. 
The process interpolates between the data distribution $p_0 = p_\text{data}$ and the prior $p_T = \B{m}$ via a monotonically decreasing noise schedule $\alpha_t \in [0,1]$:
\begin{equation}
    \label{eq:forward}
    p(\B{x}_t \mid \B{x}_0) = \text{Cat}\big(\B{x}_t; \alpha_t \B{x}_0 + (1 - \alpha_t) \B{m}\big)
\end{equation}

The sampling proceeds by simulating the reverse process, where the exact posterior is given by:
\begin{equation}
    \label{eq:reverse_posterior}
    p(\B{x}_{t-1} \mid \B{x}_t, \B{x}_0) = 
    \begin{cases} 
        \text{Cat}(\B{x}_{t-1}; \B{x}_t) & \text{if } \B{x}_t \neq \B{m}, \\[4pt] 
        \text{Cat}\left(\B{x}_{t-1}; \dfrac{(1-\alpha_{t-1}) \B{m} + (\alpha_{t-1}-\alpha_t) \B{x}_0}{1-\alpha_t}\right) & \text{if } \B{x}_t = \B{m}. 
    \end{cases}
\end{equation}
Since the clean data $\B{x}_0$ is unknown during sampling, it is approximated by a denoising neural network $\B{x}_\theta(\B{x}_t) \in \Delta^{V}$. 
Substituting this prediction into the posterior~\cref{eq:reverse_posterior} yields the parameterized reverse transition kernel $p^\text{base}(\B{x}_{t-1} \mid \B{x}_t)=p(\B{x}_{t-1} \mid \B{x}_t, \B{x}_\theta(\B{x}_t))$. 
The resulting reverse chain induces a trajectory distribution $p^{\text{base}}_{0:T}(\B{x}_{0:T})$ with time marginals $p_t^{\text{base}}(\B{x}_t)$.
The model parameters $\theta$ are optimized by minimizing a weighted cross entropy loss which is equivalent to the negative ELBO in the continuous-time limit. 
We refer to previous works~\cite{campbell2022:continuous, shi2024:simplified} for the detailed derivations. 

%% file: Sections/03_Method.tex
{\vspace{-0.25\baselineskip}}
\section{Twisted Sequential Monte Carlo for Reward Alignment}
\label{sec:twist_smc}

{\vspace{-0.25\baselineskip}}
\subsection{KL Regularized Reward Alignment}
\label{subsec:kl_reg}

Let $r : \mathcal{V}^N \to \mathbb{R}$ be a reward function that maps a fully denoised sequence of length $N$ to a scalar (\eg~human preference score or protein validity). 
Given this reward function and a pretrained model distribution $p^\text{base}_{0:T}$ parameterized by an MDM, our objective is to maximize the expected reward while penalizing deviations from the base model~\cite{Rafailov2023:DPO, jaques2017:seqtutor}:
\begin{equation}
    \label{eq:obj}
    p^*_{0:T} = \arg\max_{p_{0:T}} \mathbb{E}_{p_{0:T}}[r(\B{x}_0)] - \beta \mathcal{D}_\text{KL}(p_{0:T} \parallel p^\text{base}_{0:T}),
\end{equation}
where $\beta > 0$ is a hyperparameter controlling the strength of the KL regularization.

The intermediate target distribution $p^*_t(\B{x}_t)$ that maximizes the objective in~\cref{eq:obj} can be derived in a closed form: 
\begin{equation}
    \label{eq:target_distribution}
    p_t^*(\B{x}_t) = \frac{1}{\mathcal{Z}_t} p_t^\text{base}(\B{x}_t) \psi_t^*(\B{x}_t) \qquad \psi_t^*(\B{x}_t) = \exp\left( \frac{V_t(\B{x}_t)}{\beta} \right),
\end{equation}
where $\psi_t^*(\B{x}_t)$ is the \emph{optimal twist function} that modulates the base distribution to match the target, given as the exponentiated \emph{optimal value function} $V_t(\B{x}_t)$. 
A classic result states that the optimal value function can be expressed using the base model posterior~\cite{kappen2005:path, uehara2024:understanding, domingo:2024adjoint}:
\begin{equation}
    \label{eq:value_function}
    V_t(\B{x}_t) = \beta \log \mathbb{E}_{p^\text{base}(\B{x}_0 | \B{x}_t)}\left[ \exp\left( \frac{r(\B{x}_0)}{\beta} \right) \;\middle|\; \B{x}_t \right]. 
\end{equation}
Sampling from the target distribution in~\cref{eq:target_distribution} can be performed using Twisted Sequential Monte Carlo, whose importance weights require evaluating the optimal twist function. 
The central challenge is therefore how to estimate this twist function efficiently.
In continuous diffusion, this quantity is often approximated using Tweedie's formula~\cite{efron:2011tweedie}, which provides the posterior mean of the clean sample. This yields the plug-in estimate $r(\mathbb{E}_{p^\text{base}(\B{x}_0 \mid \B{x}_t)} [\B{x}_0])$, which has been shown to be effective in practice~\cite{Chung:2023DPS, song:2023pseudoinverse}. 
However, discrete diffusion lacks an analogous relation~\cite{raphan:2011least}, leaving costly Monte Carlo estimation as the standard practice~\cite{ou2025:smc_amot, dang:2025inference}. 
To address this challenge in discrete diffusion, we propose learning the twist function in advance to amortize this inference cost.

{\vspace{-0.25\baselineskip}}
\subsection{Twisted Sequential Monte Carlo}
\label{sec:twist SMC}
We consider the problem of sampling from the target distribution $p_t^*$ presented in~\cref{eq:target_distribution}. 
Given a proposal distribution $q(\B{x}_{t-1}|\B{x}_t)$ and the unnormalized trajectory-level target $\tilde{p}_{t:T}(\B{x}_{t:T}):=p_{t:T}^\text{base}(\B{x}_{t:T}) \psi_t^*(\B{x}_t)$, Sequential Monte Carlo (SMC)~\cite{chopin:2020introduction} interleaves sequential importance sampling with particle resampling to approximate the target distribution~\cite{del:2006sequential, Doucet2001:SMC}. 
Specifically, the unnormalized importance weight is calculated at each step as\footnote{We assume resampling at every step. See Appendix~\ref{subsec:smc_is} for a detailed discussion of SMC.}:
\begin{equation}
    \label{eq:smc_weight}
    w_{t-1} 
    = \frac{\tilde{p}_{t-1:T}(\B{x}_{t-1:T})}{\tilde{p}_{t:T}(\B{x}_{t:T}) q(\B{x}_{t-1} | \B{x}_{t})} 
    = \frac{\psi_{t-1}^*(\B{x}_{t-1})}{\psi_{t}^*(\B{x}_{t})} \frac{p^\text{base}(\B{x}_{t-1} | \B{x}_t)}{q(\B{x}_{t-1} | \B{x}_t)}, 
\end{equation}
where we adopt the Markov assumption on the target trajectory~\cite{Kim:2025DAS, ou2025:smc_amot}. 
Given $K$ particles $\{\B{x}_{t}^{(k)}\}_{k=1}^K$, the normalized weights $\{\tilde{w}_t^{(k)}\}_{k=1}^K$ yield a target approximation $p^*_t \approx \sum_{k=1}^K \tilde{w}_t^{(k)} \delta_{\B{x}_t^{(k)}}$.

The optimal proposal distribution has a closed form expression $q^*(\B{x}_{t-1} | \B{x}_t) \propto p^\text{base}(\B{x}_{t-1} | \B{x}_t) \psi_{t-1}^*(\B{x}_{t-1})$, which minimizes the variance of the importance weights in \cref{eq:smc_weight}. 
However, this optimal proposal is generally intractable, since computing its normalizing constant requires evaluating the twist function over all possible next states. 

In the cases when the reward is differentiable, one can approximate the optimal proposal via a first-order Taylor expansion~\cite{ou2025:smc_amot}:
\begin{equation}
    \tag{Grad}
    \label{eq:dg_proposal}
    q(\B{x}_{t-1} | \B{x}_t) \propto p^\text{base}(\B{x}_{t-1} | \B{x}_t) \exp \left( \B{x}_{t-1}^\top \nabla_{\B{x}} \log \psi_{t}^*(\B{x}_t) \right).
\end{equation}
This is the discrete counterpart to gradient-based guidance, an approach that has proven highly effective in the continuous domain~\cite{Chung:2023DPS, Yu:2023FreeDOM, Bansal:2023UGD}. 
However, this method exhibits two key limitations. 
First, since discrete state spaces are inherently non-differentiable, computing the gradients relies on the Gumbel-Softmax trick~\cite{jang:2016categorical}, which often suffer from gradient bias and optimization instability~\cite{paulus:2020gradient}. 
Second, and more importantly, this approach is fundamentally incompatible with non-differentiable objectives (\eg~API-based rewards).

\paragraph{Proposal Distributions.}
\label{subsec:proposal}
We consider two gradient-free alternatives that sidestep the differentiability requirement of~\cref{eq:dg_proposal}. We can either use the pretrained base transition kernel directly or fine-tune the pretrained model~\cite{wang2024:drakes, prabhudesai2024:vader,Guo2025:deepseek, zhao2025:d1} and use the resulting reward-aware proposal $p^\text{FT}(\B{x}_{t-1} | \B{x}_t)$. Each choice of transition kernel will result in the following importance weights, respectively:
\vspace{-1.25\baselineskip}
\begin{center}
\begin{minipage}{0.35\textwidth}
\begin{align}
\tag{SMC}
w_{t-1}
=
\frac{\psi_{t-1}^{*}(x_{t-1})}
     {\psi_t^{*}(x_t)}
\label{eq:base_smc}
\end{align}
\end{minipage}
\hfill
\begin{minipage}{0.6\textwidth}
\begin{align}
\tag{FT+SMC}
w_{t-1}
=
\frac{
\psi_{t-1}^{*}(x_{t-1})\,p^{\mathrm{base}}(x_{t-1}\mid x_t)
}{
\psi_t^{*}(x_t)\,p^{\mathrm{FT}}(x_{t-1}\mid x_t)
}
\label{eq:ft_smc}
\end{align}
\end{minipage}
\end{center}
\vspace{-0.5\baselineskip}
In both cases, the importance weight depends on the twist ratio $\psi_{t-1}^*/\psi_t^*$, which plays a key role in the accuracy of the target approximation.

{\vspace{-0.25\baselineskip}}
\subsubsection{Motivation: Monte Carlo Twist Function Estimation}
\label{subsec:motivation}

As discussed in \cref{subsec:kl_reg}, while the twist function can be cheaply estimated in continuous diffusion, the discrete case relies on Monte Carlo estimation. 
This approach draws $M$ clean samples from the base model posterior and averages the exponentiated rewards~\cite{ou2025:smc_amot, Li2024:SVDD, dang:2025inference}: 
\begin{equation}
    \label{eq:mc_approx}
    \psi_t^*(\B{x}_t) \approx \frac{1}{M} \sum_{m=1}^{M} \exp\left( \frac{r(\hat{\B{x}}_0^{(m)})}{\beta} \right),
    \qquad
    \{\hat{\B{x}}_0^{(m)}\}_{m=1}^M \sim p^\text{base}(\B{x}_0 | \B{x}_t).
\end{equation}
Although this estimate becomes exact as $M \to \infty$, scaling $M$ incurs significant inference overhead.

\begin{wrapfigure}{r}{0.4\textwidth}
\centering
\vspace{-1.8em}
\includegraphics[width=0.4\textwidth]{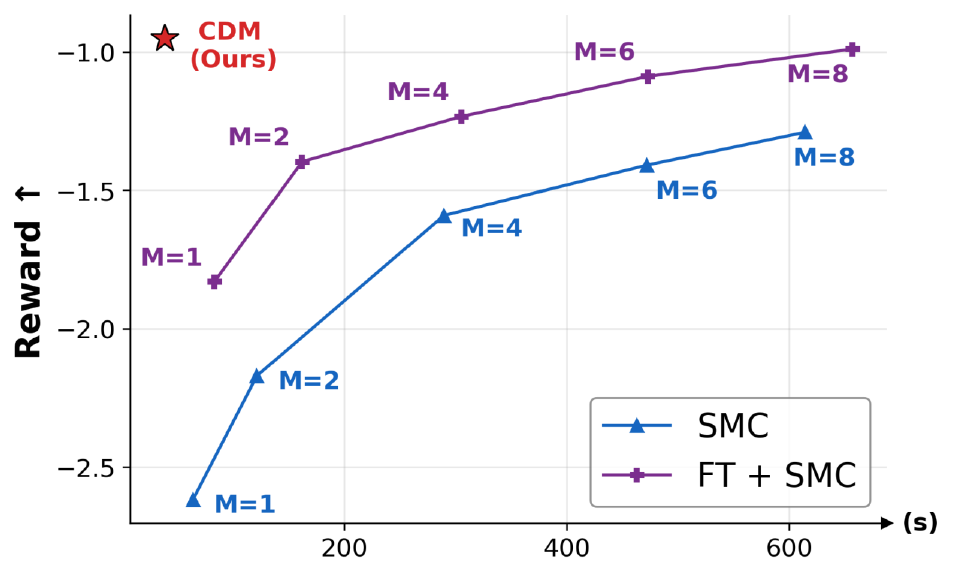}
\vspace{-1.5em}
\caption{\textbf{Reward vs. Wall-Clock Time with Varying $M$.} 
Increasing $M$ yields a more accurate twist estimate, improving SMC performance, but incurs a substantial computational cost. 
\Ours{} show superior scalability by amortizing this cost. 
}
\label{fig:monte_carlo_M}
\vspace{-2em}
\end{wrapfigure}

\Cref{fig:monte_carlo_M} illustrates the reward and wall-clock time of SMC in the protein generation task as $M$ increases. 
While a larger $M$ provides a more accurate estimate of the twist function, thereby leading to consistent improvements in reward alignment for both base and fine-tuned proposals, it also increases inference time proportionally.
This overhead becomes prohibitive when the reward evaluation is computationally expensive. 
To address this, we propose a contrastive learning framework that amortizes the twist computation by training a network to directly predict the optimal twist function $\psi_t^*$ in a single forward pass. 
This reduces the twist evaluation to a \emph{constant-time} operation and remains applicable regardless of the chosen proposal for further improvements.

{\vspace{-0.25\baselineskip}}
\section{Amortized SMC with a Learned Twist Function}
\label{sec:twist_function}
In this section, we first review the standard regression-based approach and then introduce a contrastive learning objective designed for MDMs. 

\paragraph{Regression Objective.}
Let $\psi_t^\phi(\B{x}_t)$ denote a parameterized neural network. 
The model is trained via direct regression by minimizing the Mean Squared Error between the network prediction and the optimal twist function in \cref{eq:target_distribution}:
\begin{equation}
    \label{eq:regression}
    \tag{\svdd{}}
    \min_{\phi} \mathbb{E}_{t, \B{x}_t \sim p_t^\text{base}} 
    \left[ \left( 
        \psi_t^\phi(\B{x}_t) - \mathbb{E}_{p^\text{base}(\B{x}_0 | \B{x}_t)} \left[ \exp\left(\frac{r(\B{x}_0)}{\beta}\right)\;\middle|\; \B{x}_t \right] 
    \right)^{2} \right]. 
\end{equation}
In practice, the optimal twist target is approximated via a Monte Carlo estimate with $M$ samples as in \cref{eq:mc_approx}. 
This regression target is a special case of soft Q-learning from the RL literature~\cite{Levine:2018RL, korbak:2022rl} with no intermediate reward, and has been widely adopted in prior work on diffusion models~\cite{Li2024:SVDD, Uehara:2024Finetuning}. 
For simplicity, we refer to twist-learning methods trained with this objective as \textit{\svdd{}}.
We further ablate the training-sample distribution for \svdd{} in~\cref{tab:sv_ablation}.

While straightforward, this objective trains the twist function on samples drawn from the reward-agnostic base distribution. 
Therefore, the model is trained on a distribution that does not necessarily reflect the target distribution at inference, resulting in a train-test distributional mismatch. 
As a result, the learned twist target can be inaccurate in the regions most relevant for target sampling, leading to suboptimal performance. 
This motivates a distribution-level matching objective, which naturally yields a contrastive learning formulation based on positive and negative samples.

{\vspace{-0.25\baselineskip}}
\subsection{\Ours{}: Contrastive Distribution Matching}
\label{subsec:cdm}

The core of our formulation lies in aligning the distribution induced by the twist function with the optimal target. 
Drawing inspiration from recent work on autoregressive language models~\cite{zhao2024:ctl}, we utilize the forward KL divergence. 
Specifically, let $p_t^\phi(\B{x}_t)$ denote an intermediate distribution where the base distribution is modulated by the parameterized twist function $\psi_t^\phi(\B{x}_t)$:
\begin{equation}
    \label{eq:intermediate_dist}
    p_t^\phi(\B{x}_t) = \frac{1}{\mathcal{Z}_t^\phi} p_t^{\text{base}}(\B{x}_t) \psi_t^\phi(\B{x}_t),
    \qquad
    \mathcal{Z}_t^\phi = \sum_{\B{x}_t} p_t^{\text{base}}(\B{x}_t) \psi_t^\phi(\B{x}_t). 
\end{equation}
To align $p_t^\phi$ with the optimal target $p_t^*$ at each timestep, we minimize the following time-averaged forward KL objective:
\begin{equation}
    \label{eq:cdm_obj}
    \mathcal{L}_{\text{\Ours{}}}(\phi)
    = \mathbb{E}_{t}\left[ \mathcal{D}_{\text{KL}}\left( p_t^*(\B{x}_t) \big\| p_t^\phi(\B{x}_t)\right) \right],
\end{equation}
which we refer to as Contrastive Distribution Matching, \Ours{}. 
To understand the contrastive behavior of this objective, we analyze the gradient of the loss function with respect to the parameters $\phi$:
\begin{equation}
    \label{eq:cdm_grad}
    -\nabla_\phi \mathcal{L}_{\text{\Ours{}}}(\phi)
    = \mathbb{E}_{t}\Bigg[
        \underbrace{\mathbb{E}_{p_t^*(\B{x}_t)}\left[\nabla_\phi \log \psi_t^\phi(\B{x}_t)\right]}_{\text{Positive Term}}
        -
        \underbrace{\mathbb{E}_{p_t^\phi(\B{x}_t)}\left[\nabla_\phi \log \psi_t^\phi(\B{x}_t)\right]}_{\text{Negative Term}}
    \Bigg],
\end{equation}
with the derivation deferred to Appendix~\ref{subsec:grad_derivation}. 
Note that the gradient exhibits a contrastive structure: the positive term increases $\log \psi_t^\phi$ on samples drawn from the target distribution $p_t^*$, while the negative term decreases it on samples drawn from the current approximation $p_t^\phi$. 
The positive term mitigates the train-test distributional mismatch described previously, whereas the negative term calibrates the learned distribution by suppressing suboptimal samples.
Leveraging both positive and negative samples is known to yield more effective training, as also observed in previous works~\cite{zheng:2025diffusionnft, deng:2026generative, chen:2026nft, zhu2025:DMPO}.  
Next, we explain how we effectively adapt it to diffusion models with a sampling scheme designed for efficient training.

{\vspace{-0.25\baselineskip}}
\subsection{Efficient Twist Training}
\label{subsec:pos_neg_sampling}
To evaluate the contrastive gradient, we first need to address the problem of sampling from the optimal target distribution $p_t^*$. 
Since direct sampling from the target distribution is intractable, one can approximate sampling from $p_t^*$ via importance sampling (IS) or SMC, using the pretrained base model as a proposal. 
The optimal twist function appearing in the importance weights is then estimated with the Monte Carlo approximation in~\cref{eq:mc_approx}.
While sampling from $p_t^*$ may appear circular, these samples serve as training targets that amortize the cost of all subsequent inference-time sampling. 
We analyze how the finite number of particles used for positive sampling and inference affects the sampler in Appendix~\ref{subsec:finite_particle}.

Under the IS framework, we draw trajectories from the pretrained base model and compute the importance weights only at timestep $t$, making this approach computationally efficient. 
However, in practice, we observe that it suffers from high variance when drawing positive samples.

SMC mitigates this variance by interleaving intermediate reweighting with particle resampling, yielding positive samples that are better aligned with the reward-tilted target. 
However, this improved sample quality comes at the cost of sequential weight computations, which require repeated queries to the reward model. 
This trade-off becomes especially pronounced when the reward model is expensive, motivating a more efficient training scheme. 

\paragraph{Forward-Based Gradient Estimation.}
A fundamental limitation of \cref{eq:cdm_grad} is that it allows only a \emph{single} gradient update per positive sample. 
This sample inefficiency creates a bottleneck, particularly severe when reward evaluation is computationally expensive. 
We address this by exploiting a diffusion-specific property of the target marginals. 
Rather than sampling independently from each intermediate target $p_t^*$, we first obtain clean positive samples from $p_0^*$. 
By leveraging the closed-form diffusion forward kernel, we can then draw multiple positive samples at any intermediate timestep at negligible cost. 
Specifically, the intermediate target decomposes as $p_t^*(\B{x}_t)=\sum_{\B{x}_0}p_0^*(\B{x}_0)p^{\mathrm{base}}(\B{x}_t \mid \B{x}_0)$
(Appendix~\ref{subsec:forward_kernel}), which is a structural advantage unique to diffusion frameworks and unavailable to standard autoregressive models~\cite{zhao2024:ctl, Yang:2021FUDGE, mudgal:2023controlled}. 
Leveraging this decomposition, we reformulate the gradient in \cref{eq:cdm_grad} as:
\begin{equation}
    \label{eq:cdm_grad_decomp}
    -\nabla_\phi \mathcal{L}_{\text{\Ours{}}}(\phi) = \mathbb{E}_{t}\!\left[
        \mathbb{E}_{p^*_0(\B{x}_0)} \mathbb{E}_{p^{\text{base}}(\B{x}_t \mid \B{x}_0)}\!\left[\nabla_\phi \log \psi_t^\phi(\B{x}_t)\right]
        - \mathbb{E}_{p_t^\phi(\B{x}_t)}\!\left[\nabla_\phi \log \psi_t^\phi(\B{x}_t)\right]
    \right],
\end{equation}
yielding an unbiased gradient estimator. 
This forward-based formulation enables an efficient buffer-based training scheme in which we maintain a buffer $\mathcal{B}$ of clean positive samples and repeatedly apply the forward kernel across timesteps to obtain multiple gradient updates from each sample~\cite{zhu2025:mdns, havens2025:AS}, thereby effectively reducing the cost of reward evaluations throughout training.

\vspace{-0.5\baselineskip}
\paragraph{Negative Sampling.}
One can utilize the IS/SMC framework to sample from $p_t^\phi$ by replacing the optimal twist function in \cref{eq:smc_weight} with our parameterized twist, $\psi_t^\phi$. 
Note that unlike the positive sampling case, $p_t^\phi$ does not admit a forward-kernel decomposition under the base process. 
In practice, we find that for negative sampling, IS achieves effective performance while being more computationally efficient than SMC. 
This efficiency ensures that the overall negative sampling procedure remains highly scalable.

Beyond the choice of negative sampler, we observe that purely online training of~\cref{eq:cdm_grad_decomp} exhibits optimization instability. To mitigate this, we adopt the soft target update from the RL literature~\citep{chou:2017improving, zheng:2025diffusionnft} and maintain an exponential moving average (EMA) of the twist parameters. 
The detailed training algorithm for \Ours{} is presented in Appendix~\ref{sec:algorithm}.

{\vspace{-0.25\baselineskip}}
\subsection{Efficient Twist Parameterization}
\label{subsec:efficiency}

An efficient parameterization of the twist function is critical, as we aim to amortize the expensive computational cost of Monte Carlo estimation via a single forward pass of our learned model, $\psi_t^\phi$. 
A straightforward implementation would train a separate network for $\psi_t^\phi$ from scratch, but this introduces non-negligible inference overhead, since computing the importance weights in~\cref{eq:smc_weight} requires evaluating $\psi_t^\phi$ for every particle at each denoising step.

To minimize this cost, we parameterize $\psi_t^\phi$ as a lightweight scalar head attached to the final feature layer of the pretrained model, alongside the existing logit head (\cref{fig:twist_head}). 
Thus, once the backbone features are computed, the model can produce both the logits and the twist estimate in a single forward pass through their respective heads.
This parameterization adds negligible computational overhead, around $5\%$ of the backbone forward-pass time and as little as $0.5\%$. 
As a result, sampling with the learned twist is essentially as fast as standard sampling from the base model, and the number of particles can be scaled well beyond that of the SMC baseline in~\cref{eq:base_smc}. 
This parameterization contrasts favorably with prior approaches that train isolated value networks from scratch~\citep{Li2024:SVDD, Uehara:2024Finetuning}, which not only incur non-negligible inference-time overhead but also fail to leverage the rich representations already learned by the diffusion backbone. 
We detail the twist-head architecture and parameterization in Appendix.~\ref{sec:impl_detail}.

%% file: Sections/04_Related_Works.tex
{\vspace{-0.25\baselineskip}}
\section{Related Work}
\label{sec:related_work}

\paragraph{Proposal Fine-Tuning.}
In the continuous domain, aligning diffusion models with downstream rewards typically involves either direct backpropagation across the sampling trajectory~\cite{Clark2024:DRaFT, prabhudesai2024:vader} or reformulating the denoising steps as a Markov decision process to enable reinforcement learning~\cite{Black2024:DDPO, Fan:2023DPOK, Wallace:2024DiffusionDPO}. 
While highly effective, adapting these methods to discrete state spaces necessitates specialized adaptations. 
d1~\cite{zhao2025:d1} employs a mean-field approximation to utilize the GRPO objective~\cite{shao:2024deepseekmath}, whereas DRAKES~\cite{wang2024:drakes} enables direct backpropagation through a Gumbel-Softmax relaxation~\cite{jang:2016categorical}. 
Other approaches incorporate importance sampling~\cite{zekri:2025fine, zhu2025:mdns, zhu2025:DMPO} to estimate likelihood ratios, or compute adjoint states~\cite{so:2026discrete}. 
Crucially, our amortized SMC framework is complementary to this body of work. 
We emphasize that \emph{any} fine-tuned model can be integrated as a proposal distribution within our framework to achieve further performance scaling.

\paragraph{Inference-Time Scaling.}
Inference-time scaling offers a training-free alternative, with gradient-based guidance serving as a notable example in both continuous~\cite{Chung:2023DPS} and discrete spaces~\cite{nisonoff2024:dg}. 
However, computing gradient guidance in discrete state spaces degrades when the linearity assumption is violated~\cite{zhang:2022langevin, grathwohl:2021oops}, and, more importantly, is fundamentally limited when the reward is not differentiable. 
Instead, search-based methods attempt to identify high-reward trajectories by allocating more compute for exploration during the sampling process~\cite{kim2025:rbf, ramesh:2025test, Li2024:SVDD, Ma2025:SoP, jain2026diffusion, zhang2026inference}, and have also been extended to discrete diffusion models~\cite{tang2025tr2, chen2025rfg}.
Other approaches rely on MCMC techniques, such as Gibbs sampling~\cite{chu2025:sgdd,dang:2025inference} or Metropolis-Hastings~\cite{phunyaphibarn2026:csmc}, to iteratively draw samples from the target distribution. 
However, both search- and MCMC-based methods are time-consuming and computationally heavy at inference. 
Our method addresses this by amortizing the inference cost through learning the twist function.

\paragraph{Sequential Monte Carlo for Generative Modeling.}

SMC~\cite{del:2006sequential,Doucet2001:SMC} is a widely used framework for probabilistic inference. 
Combined with continuous diffusion models, SMC is widely utilized for solving inverse problems~\cite{Wu:2023TDS, Cardoso2024:MCGDiff,trippe2022diffusion,Dou2024:FPS} and sampling from reward-tilted distributions~\cite{ren:2025driftlite, skreta:2025feynman, Kim:2025DAS, yoon:2025psi, he2025rne, he2026crepe}. 
This paradigm has been extended to discrete diffusion~\cite{lee:2025debiasing, hasan2026:dfkc,  singhal:2025general, dang:2025inference}. 
While some works consider fine-tuning the proposal distribution to amortize sampling costs~\cite{ou2025:smc_amot, holderrieth:2025leaps}, the cost of inference remains a bottleneck for expensive twist function estimation. 
In this work, we focus on amortizing this computational cost by learning the twist function, an approach that is orthogonal to proposal fine-tuning and can be combined with it for synergistic improvements.

Learning the twist functions has been studied in the context of autoregressive language modeling~\cite{Yang:2021FUDGE, mudgal:2023controlled, lawson2022sixo} and planning~\cite{lioutas:2022critic, piche:2018probabilistic}. 
A notable example is soft Q-learning from RL literature~\cite{Levine:2018RL, rawlik:2012stochastic}, which has recently been extended to the diffusion setting~\cite{Li2024:SVDD, Uehara:2024Finetuning}. 
Alternatively, \citet{zhao2024:ctl} proposed a contrastive learning objective for autoregressive language models, though their formulation is specifically designed for sequential text generation, where no forward corruption kernel is available.
We propose a contrastive objective for diffusion models that exploits the closed-form forward kernel of the diffusion process, allowing a single clean positive sample to be reused across timesteps. This yields an efficient training scheme that scales to settings with expensive reward computation, including scientific domains such as DNA and protein design.

%% file: Sections/05_Experiments.tex
{\vspace{-0.25\baselineskip}}
\section{Experiments}
\label{sec:experiments}
We evaluate \Ours{} across a diverse set of discrete diffusion reward alignment tasks. 
\Cref{subsec:applications} benchmarks our method against inference-time baselines on text generation and biological sequence design. 
\Cref{subsec:finetune_exp} integrates \Ours{} with fine-tuned proposals to evaluate their synergistic performance. 
Finally, \Cref{subsec:twist_exp} validates our contrastive objective by comparing its training convergence against \svdd{}. 
We provide implementation details and ablation studies in Appendix.~\ref{sec:impl_detail}-\ref{sec:ablation}.

{\vspace{-0.25\baselineskip}}
\subsection{Applications}
\label{subsec:applications}

\paragraph{Baselines.}
We compare \Ours{} against a representative set of scaling baselines. 
Best-of-$N$ (BoN) draws independent samples from the base model and selects the highest-reward candidate. 
We evaluate the performance of SMC under two settings, where the twist function is approximated as defined in \cref{eq:mc_approx} with $M \in \{1, 4\}$. 
\svdd{}~\cite{Li2024:SVDD, Uehara:2024Finetuning} denotes the case which learns the twist function using the regression objective defined in \cref{eq:regression}. 
For tasks with differentiable rewards, we additionally compare against SMC+Grad~\cite{ou2025:smc_amot}, which approximates the optimal proposal in via a first-order Taylor expansion~\cref{eq:dg_proposal}. 
Lastly, we include base model to serve as a reference. 

For all applications, we match training wall-clock time for methods that require optimization (\eg~\svdd{} and \Ours{}), and report scaling results against wall-clock time.
We further analyze the upfront training cost of \Ours{} and the number of downstream samples beyond which amortization pays off in Appendix~\ref{sec:cost_analysis}. 
In addition to the given reward used for scaling, we report a heldout reward that is not observed during scaling. 
We detail the experimental setup for each application below, and qualitative results are provided in Appendix~\ref{sec:qual_result}. 

\input{Figures/fig_applications}

{\vspace{-0.25\baselineskip}}
\subsubsection{Toxic Text Generation}
\label{subsec:toxic_text}

\paragraph{Experiment Setup.}
We consider the task of toxic text generation for evaluating our framework.
For the base model, we adopt a DiT-parameterized MDM~\cite{Sahoo:2024MDLM} pretrained on OpenWebText~\cite{Gokaslan2019:OpenWeb}. 
Following previous work~\cite{ou2025:smc_amot}, we use $15$ prompts for evaluation with metrics averaged over $20$ independent runs. 
The reward is the toxicity score predicted by a pretrained classifier~\cite{logacheva:2022paradetox}.
We additionally report a heldout reward computed from another classifier which was trained on multilingual dataset~\cite{dementieva:2024overview}. 

\paragraph{Results.}
Fig.~\ref{fig:applications} (a-b) show the scaling results on toxic text generation for the given and heldout reward, respectively. 
BoN scales inefficiently in this setting, consistent with observations in previous works~\cite{ou2025:smc_amot, singhal:2025general}. 
SMC+Grad incurs additional runtime from gradient backpropagation through the reward, lagging behind the other baselines in compute-matched comparisons. 
SMC exhibits more favorable scaling but is bottlenecked by the per-step Monte Carlo twist estimation, falling short of \svdd{}~\cite{Li2024:SVDD} which amortizes this cost via a learned twist head. 
\Ours{} achieves the best scaling performance, outperforming all baselines across both given and heldout reward.

{\vspace{-0.25\baselineskip}}
\subsubsection{Regulatory DNA Sequence Design}
\label{subsec:dna_design}

\paragraph{Experiment Setup.}
Following prior works~\cite{Li2024:SVDD,wang2024:drakes}, we train a CNN-based MDM~\cite{Sahoo:2024MDLM} on an enhancer activity dataset consisting of $700,000$ DNA sequences~\cite{Gosai:2023MPRA}. 
The given reward is provided by an Enformer model~\cite{Avsec:2021Enformer} trained to predict the enhancer activity in the HepG2 cell line. 
For the heldout reward, we utilize a separate Enformer model independently trained on the validation set. 

\paragraph{Results.}
\Cref{fig:applications}(c–d) show scaling performance for DNA sequence design across both given and heldout rewards. 
Similar to the toxicity task, BoN and SMC+Grad~\cite{ou2025:smc_amot} lag behind other methods as computational budget increases. 
\Ours{} maintains superior scaling, with only the SMC $M=4$ providing comparable results on the given reward, though it remains inferior on the heldout reward.

{\vspace{-0.25\baselineskip}}
\subsubsection{Protein Designability}
\label{subsec:protein_design}

\paragraph{Experiment Setup.}
We consider protein generation using DPLM-2~\cite{Wang:2025DPLM} which co-generates amino acid sequences and structures, both of which are represented as discrete tokens. 
For the generated protein to be designable, the generated amino acid sequence should fold into the generated structure. 
This is quantified via a self-consistency metric by using a folding model~\cite{Lin:2023ESM} to predict the structure of the generated sequence and computing a distance measure between the predicted and generated structure. 
The given reward is the self-consistency RMSD (\textit{scRMSD}), which uses RMSD as the distance metric. This reward is expensive to evaluate due to the folding model predictions, representing a regime where reward computation is the dominant bottleneck. 
As a heldout reward, we report \textit{scTM}, which substitutes TM-score~\cite{Zhang:2005TM} in place of RMSD.

\paragraph{Results.}
\Cref{fig:applications}(e–f) show the scaling results on the valid protein generation task for the given and heldout rewards, respectively. 
We present SMC in a separate inset, as its scaling is too slow to fit on the shared axis. 
This highlights a critical limitation of applying SMC to discrete diffusion as shown in \cref{subsec:motivation}. 
When rewards are expensive to compute, evaluating importance weights (\cref{eq:smc_weight}) introduces a severe computational bottleneck at inference. 
We observe that BoN scales comparably to \svdd{}~\cite{Li2024:SVDD}, which we assume is because the base model already assigns non-negligible density to high-reward samples. 
Notably, even in this setting with expensive reward evaluations, \Ours{} achieves superior scaling compared to both BoN and \svdd{} across the given and heldout rewards.

{\vspace{-0.25\baselineskip}}
\subsubsection{Diffusion Large Language Model Alignment}
\label{subsec:dllm_align}

\paragraph{Experiment Setup.}

We adopt LLaDA-8B-Instruct~\cite{nie:2025llada} as the base diffusion large language model (dLLM). 
We train \svdd{}~\cite{Li2024:SVDD} and \Ours{} on the training split of RewardBench~\cite{lambert:2025rewardbench}, and evaluate on the validation split. 
The given reward is the preference score from the Skywork Llama-3.1-8B model~\cite{liu:2024skywork}, which is evaluated via non-differentiable API calls and represents another computationally expensive reward scenario. 
The heldout reward is computed using the ArmoRM reward model~\cite{wang:2024interpretable}. 

\paragraph{Results.}
\Cref{fig:applications}(g–h) shows the scaling results for the dLLM alignment task across given and heldout rewards. 
As shown in the protein designability experiments in~\cref{subsec:protein_design}, SMC scales poorly due to the high cost of reward computation.
While BoN demonstrates favorable scaling, it is consistently outperformed by \Ours{}. 
This superiority is even more pronounced on the heldout reward, confirming the scalability of \Ours{} to large-scale models.

\input{Figures/fig_finetune}

{\vspace{-0.25\baselineskip}}
\subsection{Comparison with Fine-Tuned Proposals}
\label{subsec:finetune_exp}

We consider two proposal fine-tuning-based methods: d1~\cite{zhao2025:d1}, which adapts GRPO~\cite{Guo2025:deepseek} to the discrete diffusion setting, and DRAKES~\cite{wang2024:drakes}, which directly backpropagates the reward through the sampling process. 
Due to page limit, the scaling plot for DRAKES is presented in Appendix~\ref{sec:additional_results}.

\paragraph{Compatibility with Fine-Tuned Proposals.}
We evaluate the compatibility of \Ours{} with fine-tuned proposals $p^\text{FT}(\B{x}_{t-1} | \B{x}_t)$ by pairing them with the scaling methods from \cref{subsec:applications} to achieve further performance gains, while reusing the same twist head trained independently of the fine-tuned models. 
\Cref{fig:ft_combined} presents the scaling behavior of the fine-tuned proposal on the toxic text and protein generation tasks, plotted against wall-clock time. 
While all baselines gain from scaling, \Ours{} combined with fine-tuned models outperforms them in both tasks, highlighting its compatibility with fine-tuned proposals for synergistic improvements.

\paragraph{Mode Collapse of Fine-Tuned Proposals.}
Fine-tuned methods often exhibit mode collapse behaviors~\cite{Gao:2023Scaling, Kim:2025DAS} which can be particularly pronounced in applications that require high diversity, such as text or protein generation. 
In this experiment, we analyze the diversity of fine-tuned models and \Ours{} paired with the base proposal $p^{\mathrm{base}}(\B{x}_{t-1} | \B{x}_t)$ on the text and protein generation tasks. 
As diversity metrics, we report Self-BLEU~\cite{zhu:2018texygen} to evaluate text diversity by measuring the n-gram overlap ($n=4$). 
Additionally, we report generative perplexity measured with GPT2-XL~\cite{radford:2019language} to assess the quality of the generated texts. 
For protein generation, we report the number of clusters identified by FoldSeek~\cite{Van:2024FoldSeek} (Clusters) and the mean pairwise TMScore~\cite{Zhang:2005TM} (inner-TM). 
To ensure a fair comparison, we fix the inference compute budget across all methods. 
\Cref{tab:diversity} presents the diversity results for fine-tuned models compared to \Ours{}. 
In the text generation task, both d1~\cite{zhao2025:d1} and DRAKES~\cite{wang2024:drakes} exhibit mode collapse, as indicated by high Self-BLEU, along with degraded text quality, as reflected by high PPL. 
Notably, \Ours{} maintains diversity with comparable reward. 
Similarly for protein generation, d1 suffers from mode collapse, while \Ours{} mitigates mode collapse with comparable reward.

{\vspace{-0.25\baselineskip}}
\subsection{Twist Training Comparison}
\label{subsec:twist_exp}
\begin{wrapfigure}{r}{0.38\textwidth}
\centering
\vspace{-1.8em}
\includegraphics[width=0.38\textwidth]{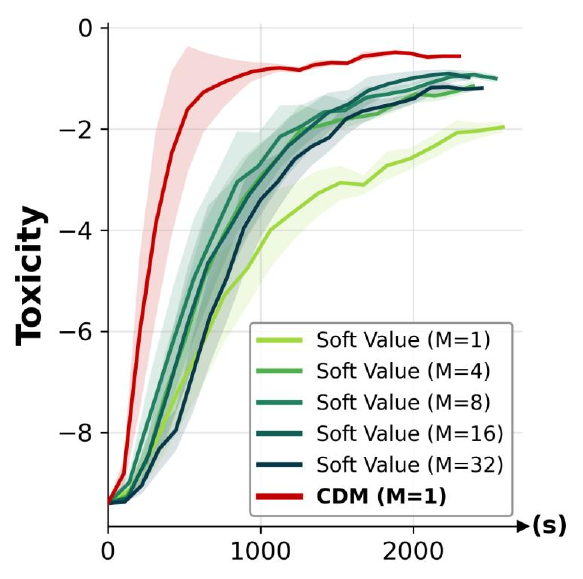}
\vspace{-1.75em}
\caption{\textbf{Training Comparison of \Ours{} with \svdd{}.}} 
\label{fig:train_convergence}
\vspace{-1.5em}
\end{wrapfigure}
In this section, we compare the training dynamics of \svdd{} and \Ours{} on toxic text and DNA sequence generation, plotting reward against wall-clock training time with fixed training parameters (\eg~optimizer, architecture, batch size). 
Additionally, for the \svdd{}~\cite{Li2024:SVDD} baseline, we sweep the Monte Carlo sample size $M$ used to estimate the optimal twist function during training. 
The training curves on toxic text generation are shown in \cref{fig:train_convergence} where \Ours{} converges faster in wall-clock time than \svdd{} with various $M$. 
We attribute this superior convergence to the use of contrastive learning with negative samples, which has been shown to be highly effective in practice~\cite{zheng:2025diffusionnft, deng:2026generative, chen:2026nft}. 
We provide the training dynamics for the regulatory DNA design task in Appendix~\ref{sec:additional_results}, which exhibit a similar pattern. 

%% file: Figures/fig_applications.tex
\begin{figure}[t]
    \centering
    \setlength{\tabcolsep}{2pt}
    \renewcommand{\arraystretch}{1.0}
    \includegraphics[width=\linewidth]{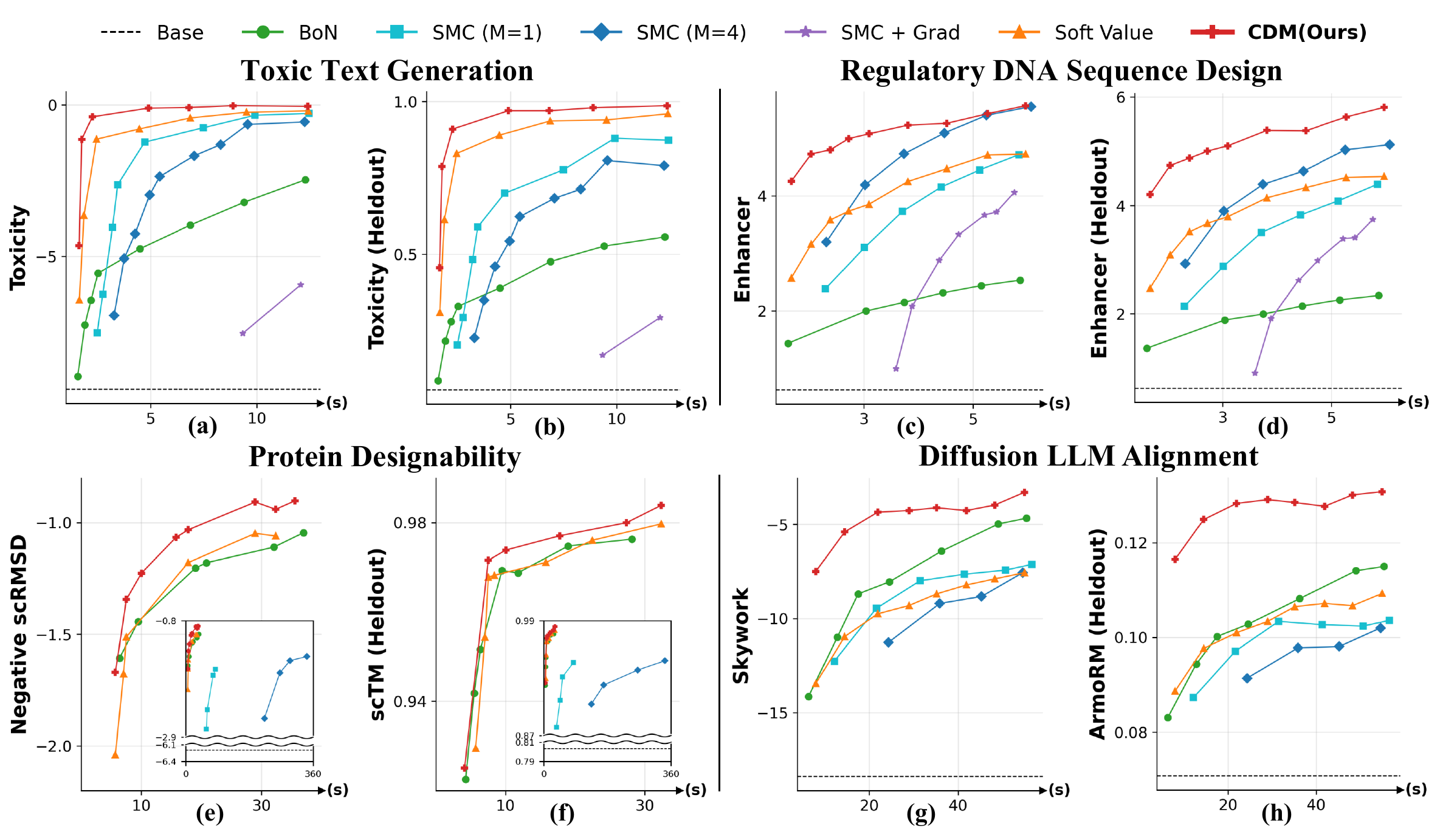} 
    \caption{
    \textbf{Scaling Results.} 
    We present scaling results for toxic text generation (a-b), regulatory DNA sequence design (c-d), protein designability (e-f), and diffusion LLM alignment (g-h). 
    For each case, we plot the given reward and a heldout reward not seen during training against inference wall-clock time. 
    In all cases, \Ours{} establishes a new Pareto front by consistently outperforming all baselines. 
    }
    \label{fig:applications}
    \vspace{-0.75\baselineskip} 
\end{figure}

%% file: Figures/fig_finetune.tex
\begin{figure}[t]
    \centering
    \begin{subfigure}[c]{0.60\textwidth} 
        \centering
        \includegraphics[width=\linewidth]{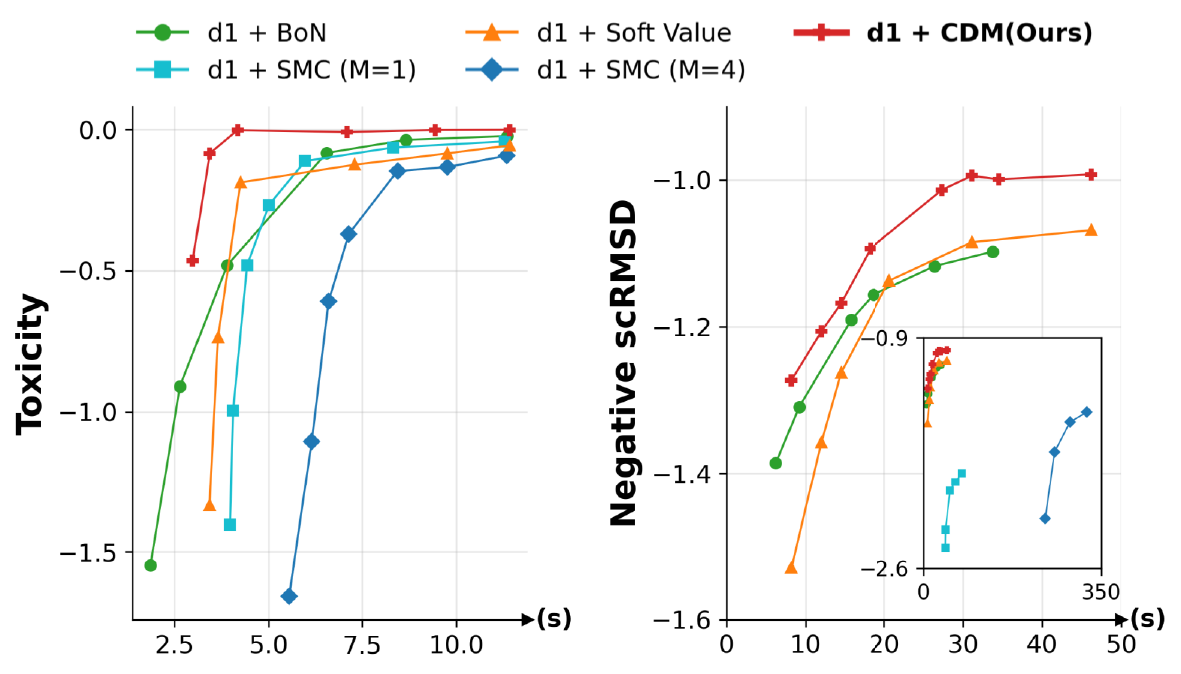} 
        \caption{Scaling with Fine-Tuned Proposals}\label{fig:ft_combined}
    \end{subfigure}\hfill
    \begin{subfigure}[c]{0.40\textwidth}
        \centering
        \renewcommand{\arraystretch}{1.4}
        
        \setlength{\tabcolsep}{3pt} 
        
        \resizebox{\linewidth}{!}{%
        \begin{tabular}{l c c c} 
            \multicolumn{4}{c}{\textbf{Toxic Text Generation}} \\ 
            \midrule
            & Reward $\uparrow$ & Self-BLEU $\downarrow$ & PPL $\downarrow$ \\
            \midrule
            d1~\cite{zhao2025:d1} & -0.933 & 0.051 & 332.322 \\
            DRAKES~\cite{wang2024:drakes} & -1.051 & 0.027 & 322.424 \\
            \rowcolor{gray!10}
            \Oursbf{} & \textbf{-0.845} & \textbf{0.015} & \textbf{124.500} \\
            \midrule 
            \addlinespace[5pt]
            \multicolumn{4}{c}{\textbf{Protein Designability}} \\
            \midrule
            & Reward $\uparrow$ & \makecell{Clusters $\uparrow$} & \makecell{Inner\\TM $\downarrow$} \\
             \midrule
            d1~\cite{zhao2025:d1} & -1.876 & 14 & 0.851 \\
            \rowcolor{gray!10}
            \Oursbf{} & \textbf{-1.723} & \textbf{17} & \textbf{0.499} \\
            \midrule
        \end{tabular}%
        } %
        
        \caption{Diversity Comparison}\label{tab:diversity}
    \end{subfigure}
    \caption{\textbf{Compatibility with Fine-Tuned Proposals.} \textbf{(Left)} Applying \Ours{} on top of fine-tuned models improves performance for both toxic text and protein generation. \textbf{(Right)} \Ours{} mitigates mode collapse commonly observed in fine-tuned models while achieving comparable rewards.}
    \label{fig:finetune}
    \vspace{-0.75\baselineskip} 
\end{figure}

%% file: Sections/06_Conclusion.tex
{\vspace{-0.25\baselineskip}}
\section{Conclusion}
\label{sec:conclusion}
\vspace{-0.5\baselineskip} 
In this work, we presented Contrastive Distribution Matching (\Ours{}), a framework that enables amortized Sequential Monte Carlo (SMC) inference in discrete diffusion by learning twist functions via a contrastive objective. 
We identified the reliance on costly Monte Carlo estimation as the primary bottleneck in twisted SMC inference and introduced a contrastive learning objective that leverages the diffusion forward process for efficient training. 
Our empirical results demonstrate that \Ours{} achieves superior scaling performance and more efficient training compared to the baselines. 
Furthermore, we showed that \Ours{} yields synergistic gains when integrated with fine-tuned proposals.

\paragraph{Limitations and Future Works.}
\Ours{} relies on positive samples drawn via SMC with Monte Carlo twist estimates, which can be poorly aligned with the target under sparse or noisy rewards, or when too few positive samples are used. 
At inference, the effective sample size can remain low when the proposal is poorly aligned with the target, which can be alleviated with fine-tuned proposals. 
Moreover, since the training cost is paid upfront, \Ours{} is most beneficial when the reward is expensive and many samples are required. 
In future work, we aim to explore more expressive twist architectures and extend our framework to binary and categorical rewards, which commonly arise in real-world applications such as agentic tasks and physical simulations.

%% file: Sections/A_Appendix.tex
\section*{Societal Impacts}
In this work, we adapt \Ours{} to sample from a reward-tilted distribution to optimize downstream objectives. 
We evaluate our method on toxic text generation strictly for safety benchmarking. 
Notably, this same technique can be inverted via a negative reward to actively reduce toxicity and mitigate harmful model behaviors. 
Furthermore, our application to protein generation holds the potential to accelerate therapeutic development and drug discovery. 
Lastly, applying our method to LLM alignment enhances the reliability and helpfulness of model interactions for end users.

\input{Sections/A3_Related_Work}
\input{Sections/A1_Derivations}

\input{Sections/A2_CDM_Algorithm}

\input{Sections/A3_Implementation_Details}

\input{Sections/A4_Additional_Results}

\clearpage
\newpage

%% file: Sections/A1_Derivations.tex
\section{Derivations}
\label{sec:proof}

We consider the problem of sampling from the intermediate target distribution $p_t^*(\B{x}_t)=\frac{1}{\mathcal{Z}_t}p_t^\text{base}(\B{x}_t)\psi_t^*(\B{x}_t)$, and denote its unnormalized density by $\tilde{p}_t^*(\B{x}_t)=p_t^\text{base}(\B{x}_t)\psi_t^*(\B{x}_t)$.
For use in importance sampling (IS) and Sequential Monte Carlo (SMC), we lift this marginal target to a trajectory-level target.
Let $\tilde{\pi}_{t:T}(\B{x}_{t:T})$ be an arbitrary unnormalized path-space extension whose marginal over future variables recovers the unnormalized target at time $t$:
\begin{align}
\nonumber
\sum_{\B{x}_{t+1:T}}\tilde{\pi}_{t:T}(\B{x}_{t:T})=\tilde{p}_t^*(\B{x}_t).
\end{align}
Equivalently, after normalization, the corresponding trajectory-level distribution has time-$t$ marginal $p_t^*(\B{x}_t)$.
This formulation allows us to apply IS and SMC by sampling trajectories $\B{x}_{t:T}$ from a proposal distribution and using their time-$t$ components to approximate the desired marginal target.

\subsection{Importance Sampling and Sequential Monte Carlo}
\label{subsec:smc_is}

\paragraph{Importance Sampling.}
Given a proposal distribution $q_{t:T}(\mathbf{x}_{t:T})$, IS corrects for the distributional mismatch with the target distribution $p^*_{t:T}$ using importance weights. 
The unnormalized importance weights $\{{W}_t \}_{k=1}^K$ and the self-normalized importance weights $\{ \tilde{W}_t \}_{k=1}^K$ are computed as:
\begin{equation}
    W_t^{(k)} = \frac{\tilde{\pi}_{t:T}(\B{x}_{t:T}^{(k)})}{q_{t:T}(\B{x}_{t:T}^{(k)})}, \qquad \tilde{W}_t^{(k)} = \frac{W_t^{(k)}}{\sum_{j=1}^K W_t^{(j)}}
\end{equation} 
The normalized weights $\{ \tilde{W}_t \}_{k=1}^K$ define an empirical distribution that enables approximate sampling from the target. 
However, IS often suffers from weight degeneracy problem and results in high-variance estimators.

\paragraph{Sequential Monte Carlo.}

To mitigate weight degeneracy, SMC interleaves sequential importance sampling with particle resampling. When extending a trajectory from $\B{x}_{t:T}$ to $\B{x}_{t-1:T}$, the incremental importance weight $w_{t-1}$ is computed as:
\begin{equation}
    \label{eq:weight_traj}
    w_{t-1} = \frac{\tilde{\pi}_{t-1:T}(\B{x}_{t-1:T})}{\tilde{\pi}_{t:T}(\B{x}_{t:T})\, q(\B{x}_{t-1} \mid \B{x}_t)}.
\end{equation}
The unnormalized importance weight is then updated by multiplying the previous weight by an incremental importance weight: $W_{t-1}=W_t\cdot w_{t-1}$.

Under a Markovian assumption on the target trajectory, the joint distributions factorize as
\begin{align}
    \label{eq:traj_target_tminus1}
    \tilde{\pi}_{t-1:T}(\B{x}_{t-1:T})
    &= \tilde{p}_{t-1}(\B{x}_{t-1}) \prod_{s=t}^{T} \gamma_s(\B{x}_s \mid \B{x}_{s-1}), \\
    \label{eq:traj_target_t}
    \tilde{\pi}_{t:T}(\B{x}_{t:T})
    &= \tilde{p}_t(\B{x}_t) \prod_{s=t+1}^{T} \gamma_s(\B{x}_s \mid \B{x}_{s-1}),
\end{align}
where $\gamma_s(\B{x}_s \mid \B{x}_{s-1})$ is an arbitrary forward kernel.

Substituting~\cref{eq:traj_target_tminus1,eq:traj_target_t} into~\cref{eq:weight_traj}, the product over $s = t+1, \dots, T$ cancels between numerator and denominator, leaving only the $s = t$ factor from the numerator:
\begin{equation}
    \label{eq:weight_general}
    w_{t-1}
    = \frac{\tilde{p}_{t-1}(\B{x}_{t-1}) \prod_{s=t}^{T} \gamma_s(\B{x}_s \mid \B{x}_{s-1})}{\tilde{p}_t(\B{x}_t) \prod_{s=t+1}^{T} \gamma_s(\B{x}_s \mid \B{x}_{s-1})\, q(\B{x}_{t-1} \mid \B{x}_t)}
    = \frac{\tilde{p}_{t-1}(\B{x}_{t-1})}{\tilde{p}_t(\B{x}_t)}\, \frac{\gamma_t(\B{x}_t \mid \B{x}_{t-1})}{q(\B{x}_{t-1} \mid \B{x}_t)}.
\end{equation}

Specifically, we choose the forward kernel to be the diffusion model's forward kernel~\cite{Kim:2025DAS, ou2025:smc_amot}:
\begin{equation}
    \label{eq:bayes_kernel}
    \gamma_t(\B{x}_t \mid \B{x}_{t-1}) \coloneqq p^{\text{base}}(\B{x}_t|\B{x}_{t-1})=\frac{p^\text{base}(\B{x}_{t-1} \mid \B{x}_t)\, p^\text{base}_t(\B{x}_t)}{p^\text{base}_{t-1}(\B{x}_{t-1})}.
\end{equation}
Under this choice, the unnormalized trajectory-level target reduces to
$p_{t:T}^\text{base}(\B{x}_{t:T})\psi_t^*(\B{x}_t)$, 
which is the form used in~\cref{eq:smc_weight}.

Substituting~\cref{eq:bayes_kernel} into~\cref{eq:weight_general} gives:
\begin{equation}
    \label{eq:weight_expanded}
    w_{t-1}
    = 
      \frac{p^\text{base}_{t-1}(\B{x}_{t-1})\, \psi^*_{t-1}(\B{x}_{t-1})}{p^\text{base}_t(\B{x}_t)\, \psi^*_t(\B{x}_t)}\,
      \frac{p^\text{base}(\B{x}_{t-1} \mid \B{x}_t)\, p^\text{base}_t(\B{x}_t)}{p^\text{base}_{t-1}(\B{x}_{t-1})\, q(\B{x}_{t-1} \mid \B{x}_t)},
\end{equation}
where the base marginals $p^\text{base}_{t-1}(\B{x}_{t-1})$ and $p^\text{base}_t(\B{x}_t)$ cancel out, leading to,
\begin{equation}
    w_{t-1}
    = \frac{\psi^*_{t-1}(\B{x}_{t-1})}{\psi^*_t(\B{x}_t)}\,
            \frac{p^\text{base}(\B{x}_{t-1} \mid \B{x}_t)}{q(\B{x}_{t-1} \mid \B{x}_t)}.
\end{equation}

After computing incremental weights, SMC resamples particles by drawing ancestor indices from the categorical distribution defined by the normalized weights:
\begin{equation}
    a_{t-1}^{(k)} \sim \mathrm{Cat}\!\left( \left\{ \tilde{W}_{t-1}^{(i)} \right\}_{i=1}^{K} \right),
    \qquad
    \tilde{W}_{t-1}^{(i)} = \frac{W_{t-1}^{(i)}}{\sum_{j=1}^{K} W_{t-1}^{(j)}},
\end{equation}
and replaces each particle by its sampled ancestor, $\B{x}_{t-1}^{(k)} \leftarrow \B{x}_{t-1}^{(a_{t-1}^{(k)})}$. 
In practice, resampling is triggered adaptively based on the effective sample size $\mathrm{ESS} = 1 / \sum_{k=1}^{K} (\tilde{W}_t^{(k)})^2$, with resampling performed whenever ESS falls below a threshold~\cite{chopin:2020introduction}. After resampling, the unnormalized weights are reset to
$W_{t-1}^{(k)}=1$ for all particles.
Setting $\mathrm{ESS}_\text{thres} = 0$ disables resampling at every step, reducing SMC to IS. 
We present the full inference procedure in~\cref{alg:smc}.

The resulting particle system yields asymptotically consistent estimators of expectations under the intermediate target $p_t^*$, together with an empirical approximation of $p_t^*$ itself:
\begin{equation}
    \label{eq:smc_estimator}
    \mathbb{E}_{p_t^*}\!\left[f(\B{x}_t)\right] \approx \sum_{k=1}^{K} \tilde{W}_t^{(k)}\, f(\B{x}_t^{(k)}),
    \qquad
    p_t^* \approx \sum_{k=1}^{K} \tilde{W}_t^{(k)}\, \delta_{\B{x}_t^{(k)}}.
\end{equation}

\subsection{Contrastive Gradient Derivations}
\label{subsec:grad_derivation}

We provide the derivation of the contrastive gradient~\cref{eq:cdm_grad} for completeness. 
A formulation within the autoregressive language model is demonstrated in~\cite{zhao2024:ctl}. 

Plugging the parameterized tilted distribution from~\cref{eq:intermediate_dist} into the forward KL objective yields:
\begin{align}
    \mathcal{D}_{\text{KL}}\!\left(p_t^*(\B{x}_t) \,\big\|\, p_t^\phi(\B{x}_t)\right)
    &= \mathbb{E}_{p_t^*(\B{x}_t)}\!\left[\log p_t^*(\B{x}_t) - \log p_t^\phi(\B{x}_t)\right] \\
    &= \mathbb{E}_{p_t^*(\B{x}_t)}\!\left[\log p_t^*(\B{x}_t) - \log p_t^{\text{base}}(\B{x}_t) - \log \psi_t^\phi(\B{x}_t)\right] + \log \mathcal{Z}_t^\phi.
\end{align}

Noting that the first two terms in the expectation are invariant to $\phi$, the gradient with respect to the parameters $\phi$ simplifies to: 
\begin{equation}
    \label{eq:kl_grad_intermediate}
    \nabla_\phi\, \mathcal{D}_{\text{KL}}\!\left(p_t^* \,\big\|\, p_t^\phi\right)
    = -\,\mathbb{E}_{p_t^*(\B{x}_t)}\!\left[\nabla_\phi \log \psi_t^\phi(\B{x}_t)\right] + \nabla_\phi \log \mathcal{Z}_t^\phi.
\end{equation}

With log-derivative trick, the normalization constant expands as follows:
\begin{align}
    \nabla_\phi \log \mathcal{Z}_t^\phi 
    &= \frac{1}{\mathcal{Z}_t^\phi}\, \nabla_\phi \sum_{\B{x}_t} p_t^{\text{base}}(\B{x}_t)\, \psi_t^\phi(\B{x}_t) \\
    &= \sum_{\B{x}_t} \frac{p_t^{\text{base}}(\B{x}_t)\, \psi_t^\phi(\B{x}_t)}{\mathcal{Z}_t^\phi}\, \nabla_\phi \log \psi_t^\phi(\B{x}_t)
    = \mathbb{E}_{p_t^\phi(\B{x}_t)}\!\left[\nabla_\phi \log \psi_t^\phi(\B{x}_t)\right],
\end{align}
where the last equality uses the definition of $p_t^\phi$ in~\cref{eq:intermediate_dist}.

Substituting this result into~\cref{eq:kl_grad_intermediate} and taking the expectation over time $t$ yields the contrastive gradient in~\cref{eq:cdm_grad}:
\begin{equation} 
    -\nabla_\phi\, \mathcal{L}_{\text{\Ours{}}}(\phi)
    = \mathbb{E}_{t}\!\left[
    \mathbb{E}_{p_t^*(\B{x}_t)}\!\left[\nabla_\phi \log \psi_t^\phi(\B{x}_t)\right]
    - \mathbb{E}_{p_t^\phi(\B{x}_t)}\!\left[\nabla_\phi \log \psi_t^\phi(\B{x}_t)\right]
    \right].
\end{equation}

\subsection{Forward-Based Gradient Estimation.}
\label{subsec:forward_kernel}

In this section, we show that the positive-term gradient estimator used in~\cref{eq:cdm_grad_decomp},
\begin{equation}
    \mathbb{E}_{p^*_0(\B{x}_0)}\, \mathbb{E}_{p^{\text{base}}(\B{x}_t \mid \B{x}_0)}\!\left[\nabla_\phi \log \psi_t^\phi(\B{x}_t)\right],
\end{equation}
is identical to the marginal expectation
\begin{equation}
    \mathbb{E}_{p_t^*(\B{x}_t)}\!\left[\nabla_\phi \log \psi_t^\phi(\B{x}_t)\right]. 
\end{equation}

The KL-regularized objective in~\cref{eq:obj} admits the following closed-form solution over joint trajectory distributions~\citep{Uehara:2024Finetuning, uehara2024:understanding}:
\begin{equation}
    \label{eq:traj_opt}
    p^*_{0:T}(\B{x}_{0:T}) = \frac{1}{\mathcal{Z}}\, p^{\text{base}}_{0:T}(\B{x}_{0:T})\, \exp\!\left(\frac{r(\B{x}_0)}{\beta}\right), \qquad \mathcal{Z} = \mathbb{E}_{p^{\text{base}}_0(\B{x}_0)}\!\left[\exp(r(\B{x}_0)/\beta)\right].
\end{equation}

This closed-form solution follows from a standard variational argument.
For any trajectory distribution $q_{0:T}$, consider the functional $\mathcal{J}$ given as:
\begin{align}
    \mathcal{J}(q_{0:T})=\sum_{\B{x}_{0:T}}q_{0:T}(\B{x}_{0:T}) r(\B{x}_0)-\beta\sum_{\B{x}_{0:T}}q_{0:T}(\B{x}_{0:T})\log
    \frac{q_{0:T}(\B{x}_{0:T})}{p^{\text{base}}_{0:T}(\B{x}_{0:T})}.
\end{align}
Introducing a Lagrange multiplier $\lambda$ for the constraint
$\sum_{\B{x}_{0:T}}q_{0:T}(\B{x}_{0:T})=1$, the stationarity condition gives
\begin{align}
    r(\B{x}_0)-\beta\left(\log\frac{q_{0:T}(\B{x}_{0:T})}{p^{\text{base}}_{0:T}(\B{x}_{0:T})}+1\right)+\lambda=0.
\end{align}
Rearranging yields $q_{0:T}(\B{x}_{0:T})\propto p^{\text{base}}_{0:T}(\B{x}_{0:T})\exp\left(\frac{r(\B{x}_0)}{\beta}\right)$.
Normalizing this density gives the unique optimizer of the KL-regularized objective:
\begin{align}
    \label{eq:traj_opt3}
    p^*_{0:T}(\B{x}_{0:T})=\frac{1}{\mathcal{Z}}p^{\text{base}}_{0:T}(\B{x}_{0:T})\exp\left(\frac{r(\B{x}_0)}{\beta}\right),
\end{align}
where
\begin{align}
    \mathcal{Z}=\sum_{\B{x}_{0:T}}p^{\text{base}}_{0:T}(\B{x}_{0:T})\exp\left(\frac{r(\B{x}_0)}{\beta}\right)=
    \mathbb{E}_{p^{\text{base}}_0(\B{x}_0)}
    \left[\exp\left(\frac{r(\B{x}_0)}{\beta}\right)\right].
\end{align}

Decomposing $p^{\text{base}}_{0:T}(\B{x}_{0:T}) = p^{\text{base}}_0(\B{x}_0)\, p^{\text{base}}(\B{x}_{1:T} \mid \B{x}_0)$ and collecting the reward-dependent terms into the clean marginal in~\cref{eq:traj_opt3}, we obtain: 
\begin{equation}
    \label{eq:traj_factored}
    p^*_{0:T}(\B{x}_{0:T}) = \underbrace{\frac{p^{\text{base}}_0(\B{x}_0)\, \exp(r(\B{x}_0)/\beta)}{\mathcal{Z}}}_{=\, p^*_0(\B{x}_0)}\; p^{\text{base}}(\B{x}_{1:T} \mid \B{x}_0).
\end{equation}
Since $p^*_{0:T}(\B{x}_{0:T})$ also factorizes as
$p^*_0(\B{x}_0)p^*(\B{x}_{1:T} \mid \B{x}_0)$, comparing the two expressions gives
\begin{align}
    \label{eq:traj_factored2}
    p^{*}(\B{x}_{1:T} \mid \B{x}_0)
    =
    p^{\text{base}}(\B{x}_{1:T} \mid \B{x}_0).
\end{align}

Marginalizing~\cref{eq:traj_factored2} over $(\B{x}_{1:t-1}, \B{x}_{t+1:T})$ for any $t \geq 1$ gives the per-step forward marginal:
\begin{equation}
    \label{eq:marginal_kernel_eq}
    \begin{aligned}
        p^*(\B{x}_t \mid \B{x}_0)
        = \sum_{\substack{\B{x}_{1:t-1},\\ \B{x}_{t+1:T}}} p^*(\B{x}_{1:T} \mid \B{x}_0) = \sum_{\substack{\B{x}_{1:t-1},\\ \B{x}_{t+1:T}}} p^{\text{base}}(\B{x}_{1:T} \mid \B{x}_0) = p^{\text{base}}(\B{x}_t \mid \B{x}_0).
    \end{aligned}
\end{equation}

Using~\cref{eq:marginal_kernel_eq}, the target joint marginal over $(\B{x}_0,\B{x}_t)$ factorizes as
$p^*_{0,t}(\B{x}_0,\B{x}_t)=p^*_0(\B{x}_0)\,p^{\text{base}}(\B{x}_t \mid \B{x}_0)$.
Therefore, marginalizing over $\B{x}_0$ yields
\begin{equation}
    \label{eq:marginal_target_factored}
    p_t^*(\B{x}_t)
    =
    \sum_{\B{x}_0}
    p^*_0(\B{x}_0)\,
    p^{\text{base}}(\B{x}_t \mid \B{x}_0).
\end{equation}

Substituting~\cref{eq:marginal_target_factored} into the positive term of~\cref{eq:cdm_grad} yields:
\begin{align}
    \mathbb{E}_{p_t^*(\B{x}_t)}\!\left[\nabla_\phi \log \psi_t^\phi(\B{x}_t)\right]
    &= \sum_{\B{x}_t} p_t^*(\B{x}_t)\, \nabla_\phi \log \psi_t^\phi(\B{x}_t) \\
    &= \sum_{\B{x}_t} \sum_{\B{x}_0} p^*_0(\B{x}_0)\, p^{\text{base}}(\B{x}_t \mid \B{x}_0)\, \nabla_\phi \log \psi_t^\phi(\B{x}_t) \\
    &= \mathbb{E}_{p^*_0(\B{x}_0)}\, \mathbb{E}_{p^{\text{base}}(\B{x}_t \mid \B{x}_0)}\!\left[\nabla_\phi \log \psi_t^\phi(\B{x}_t)\right].
\end{align}
This establishes the forward-based reformulation of the positive term used in~\cref{eq:cdm_grad_decomp}.

\subsection{Effect of Finite Particle Approximation}
\label{subsec:finite_particle}

We provide a theoretical analysis of the bias and variance due to the finite particle approximation. 
We distinguish three effects: (1) using the exact terminal reward potential preserves the target distribution regardless of the intermediate learned twist, (2) finite-particle self-normalized estimates generally have bias and variance, while remaining consistent as $K_\text{inf}\to\infty$, and (3) finite positive-sample budgets can affect the quality of the learned twist and hence the sampling variance. We give an informal argument for the third effect.

\paragraph{Exact Terminal Weighting Preserves the Target at Inference.}
While the terminal particle approximation is consistent~\cite{zhao2024:ctl}, the empirical distribution is generally biased for finite $K_\text{inf}$.
Following~\citet{zhao2024:ctl}, we use the learned twist function $\psi_t^\phi$ at intermediate timesteps and the exact reward potential only at the terminal timestep. 
The unnormalized intermediate targets are given as:
\begin{equation}
    \tilde{p}_t(\B{x}_t) = 
    \begin{cases}
        p_t^\text{base}(\B{x}_t)\, \psi_t^\phi(\B{x}_t), & t \in \{1, \dots, T\}, \\[4pt]
        p_0^\text{base}(\B{x}_0) \exp\left(r(\B{x}_0)/\beta\right), & t = 0.
    \end{cases}
\end{equation}
We write the exact terminal potential as $G(\B{x}_0) \coloneqq \exp(r(\B{x}_0)/\beta)$. 
The incremental weight is then given as:
\begin{equation}
    w_{t-1} = \frac{p^\text{base}(\B{x}_{t-1} \mid \B{x}_t)}{q(\B{x}_{t-1} \mid \B{x}_t)} \frac{\Psi_{t-1}^\phi(\B{x}_{t-1})}{\Psi_t^\phi(\B{x}_t)}, 
    \qquad 
    \Psi_0^\phi = G, \quad \Psi_t^\phi = \psi_t^\phi \text{ for } t > 0.
\end{equation}
After multiplying the incremental weights along a trajectory, all the twist factors $\Psi_{t-1}^\phi(\B{x}_{t-1}) / \Psi_t^\phi(\B{x}_t)$ telescope, leaving only the terminal reward. 
Therefore, the intermediate twist does not bias the final unnormalized target, but mainly affects the sampling variance.

\paragraph{Finite-Particle Approximation During Inference Affects Sampling Variance.}
Let $\hat{p}_t^{K_\text{inf}}$ be the empirical distribution obtained from SMC sampling with $K_\text{inf}$ particles at inference. 
By the central limit theorem (CLT) for SMC~\cite{chopin:2004central}, we have:
\begin{equation}
    \sqrt{K_\text{inf}} \left( \hat{p}_t^{K_\text{inf}}(\B{x}_t) - p_t^{\Psi_t}(\B{x}_t) \right) \rightarrow \mathcal{N}\left(0, \sigma_t^2(\Psi_t)\right),
\end{equation}
which gives the rate $\mathrm{Var}[\hat{p}_t^{K_\text{inf}}] = O(\sigma_t^2(\Psi_t)/K_\text{inf})$~\cite{doucet:2009tutorial}.
Thus, as the number of particles during inference increases, the variance of the approximate distribution decreases asymptotically.
Furthermore, since $\Psi_0 = G$, increasing the number of particles ensures convergence to the correct terminal target distribution.
Note that this holds for any fixed twist.

\paragraph{The Learned Twist Error Affects the Sampling Variance at Inference.}
We now give an informal argument for how the finite training budget $K_\text{train}$ propagates to inference. 
Note that this analysis is meant to convey the qualitative behavior rather than a tight guarantee. 
The twist $\psi_t^\phi$ is itself learned from a finite number of positive-sample particles. 
The positive sampler in \Ours{} training (\cref{alg:cdm_training}) is an SMC procedure with $K_\text{train}$ particles targeting $p_0^*$, so by the same SMC CLT applied at training time, its empirical target converges at rate $K_\text{train}^{-1/2}$. 
Under the ideal-training assumption, the learned twist is a smooth functional of this empirical target whose population limit recovers the optimal twist $\psi_t^*$, and thus inherits the same rate. 
Writing the log-twist error as:
\begin{equation}
    \varepsilon_t(\B{x}) \coloneqq \log \psi_t^*(\B{x}) - \log \psi_t^\phi(\B{x}),
\end{equation}
we assume that it vanishes at the training rate, $\varepsilon_t = O(K_\text{train}^{-1/2})$. 
This error is the only quantity through which the finite training budget reaches inference. 

Now we fix such a twist and run inference SMC with $K_\text{inf}$ particles. 
Since the terminal potential is exact, the terminal sampler stays consistent, and the error $\varepsilon_t$ perturbs the CLT variance constant, which we write as $\sigma_t^2 = \sigma_t^2(\varepsilon_t)$ with $\sigma_t^2(0) = \sigma_t^{*2}$ at the optimal twist.
Assuming that the variance depends smoothly on the twist, since $\varepsilon_t = O(K_\text{train}^{-1/2})$, the variance constant is perturbed at the following rate:
\begin{equation}
    \sigma_t^2(\varepsilon_t) = \sigma_t^{*2} + O\left(K_\text{train}^{-1/2}\right).
\end{equation}
Combining with the previous result, we have the following informal expression:
\begin{equation}
    \mathrm{Var}\left[\hat{p}_t^{K_\text{inf}}\right] 
    = \frac{\sigma_t^{*2}}{K_\text{inf}} \left(1 + O\left(K_\text{train}^{-1/2}\right)\right)
    = \frac{\sigma_t^{*2}}{K_\text{inf}} + O\left(\frac{1}{K_\text{inf} \sqrt{K_\text{train}}}\right).
\end{equation}
At an intuitive level, this suggests that with the learned twist, the empirical terminal distribution stays consistent and its variance vanishes at rate $1/K_\text{inf}$, while $K_\text{train}$ controls the size of the second term.
Furthermore, since SMC is sequential, its asymptotic variance depends on the full sequence of twists.
As this derivation is non-trivial and beyond the scope of our paper, we leave this sequential analysis of the variance to future work. 
In practice, additional factors such as optimization convergence, model capacity, and approximation error may also influence the learned twist, so the effective role of $K_\text{train}$ can be more nuanced. 
We present the corresponding empirical results in~\cref{tab:particle_size}.

%% file: Sections/A2_CDM_Algorithm.tex
\section{Algorithms}
\label{sec:algorithm}
In this section, we present the algorithmic implementation of Twisted Sequential Monte Carlo (SMC) and the complete \Ours{} training loop. The detailed pseudocode for both procedures is provided in~\cref{alg:smc,alg:cdm_training}.

\subsection{Twisted SMC}
\label{sec:algorithm_smc}

As discussed in \cref{subsec:smc_is}, the algorithm interleaves sequential importance sampling with particle resampling based on the importance weights. 
To establish a unified algorithmic framework, \cref{alg:smc} operates on a generic twist function, denoted as $\Psi$. 

For positive sampling, $\Psi$ is instantiated as the Monte Carlo estimate $\hat{\psi}$ defined in~\cref{eq:mc_approx} to approximate the target distribution $p^*$.  
For the negative sampling, $\Psi$ is instantiated as the parameterized twist network $\psi^\phi$ to sample from $p^\phi$.
At inference, we use $\Psi_t=\psi_t^\phi$ for $t>0$ and the exact terminal potential $\Psi_0(\B{x}_0)=\exp(r(\B{x}_0)/\beta)$, as in Appendix~\ref{subsec:finite_particle}.
Furthermore, the resampling behavior is governed by an Effective Sample Size (ESS) threshold, $\mathrm{ESS}_\mathrm{thres}$.
Notably, by setting $\mathrm{ESS}_\mathrm{thres} = 0$, we explicitly disable the resampling mechanism, reducing SMC to standard Importance Sampling (IS). Note that for IS,  the twist ratios $\Psi_{t-1}(\B{x}_{t-1}^{(k)}) / \Psi_t(\B{x}_t^{(k)})$ telescope across timesteps and therefore do not need to be computed explicitly except at the final step $t_\text{stop}$. In particular, for $t-1 > t_{\text{stop}}$, we may set $w_{t-1}^{(k)}=p^\text{base}(\B{x}_{t-1}^{(k)} \mid \B{x}_t^{(k)}) / q(\B{x}_{t-1}^{(k)} \mid \B{x}_t^{(k)})$ and calculate the twist function only at the final step via $w_{t_\text{stop}}^{(k)} = \Psi_{t_\text{stop}}(\B{x}_{t_\text{stop}}^{(k)})p^\text{base}(\B{x}_{t_\text{stop}}^{(k)} \mid \B{x}_{t_\text{stop}+1}^{(k)}) / q(\B{x}_{t_\text{stop}}^{(k)} \mid \B{x}_{t_\text{stop}+1}^{(k)})$, where we omitted $\Psi_{T}(\B{x}_{T}^{(k)})$ since all $\{\B{x}_{T}^{(k)}\}_{k=1}^K$ are at the same mask state $\B{m}$.
\Cref{fig:concept_figure} compares standard SMC with the proposed amortized inference approach using a learned twist. 

\input{Figures/concept_figure}
\input{Algorithms/smc}

\input{Algorithms/cdm}

\subsection{\Ours{}: Twist Training}
\label{sec:algorithm_cdm}

Building on this SMC framework, \cref{alg:cdm_training} outlines the full training procedure for Contrastive Distribution Matching (\Ours{}), which estimates the contrastive gradient in~\cref{eq:cdm_grad_decomp} efficiently by exploiting the diffusion forward process. 

For positive sampling, we draw clean samples via SMC and store them in a buffer $\mathcal{B}$ of capacity $B_\text{buffer}$, refreshed every $n_\text{update}$ steps. 
We then draw $\B{x}_0^* \sim \mathcal{B}$ and apply the cheap forward kernel~\cref{eq:forward} to obtain $\B{x}_t^*$, reusing each clean sample across many gradient updates.
For the negative term, we draw $\B{x}_t^\phi$ via importance sampling under the EMA-updated twist $\psi^{\phi_\text{EMA}}$. 
We present implementation details and ablations of the key hyperparameters in the next section.

%% file: Figures/concept_figure.tex
\begin{figure}[t!]
    \centering
    \includegraphics[width=0.95\linewidth]{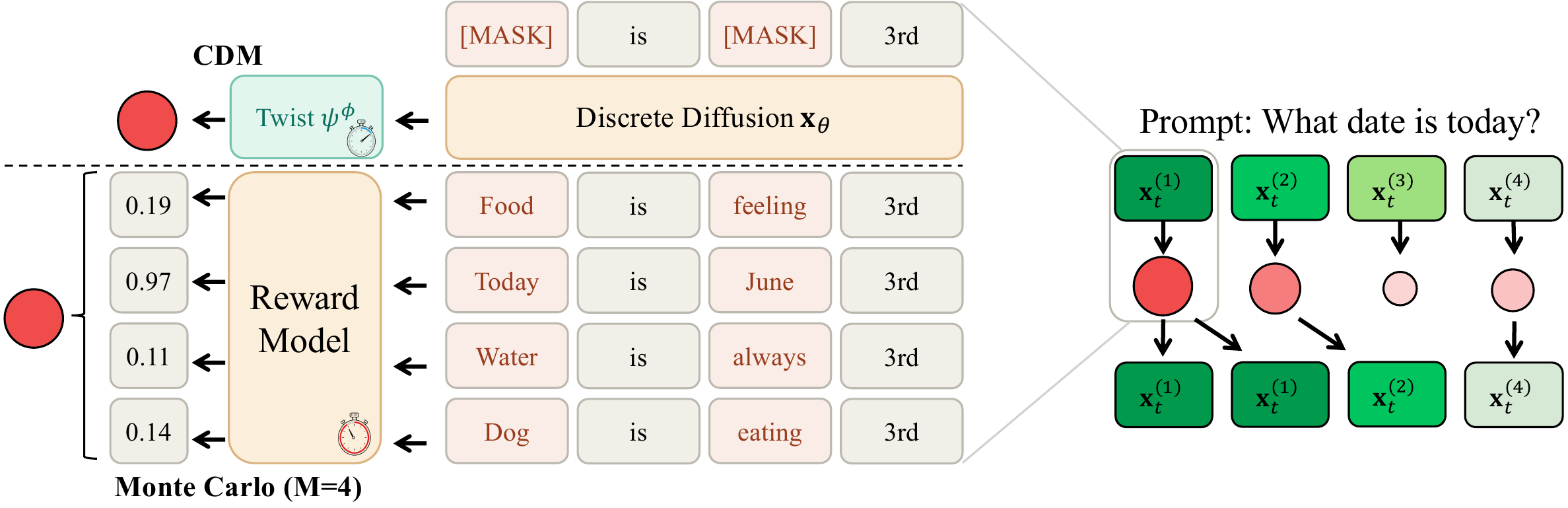}
    \caption{
    \textbf{Amortized Twisted SMC Procedure.}
    With the learned twist function, we can amortize the SMC inference with a single forward pass. 
    On the other hand, SMC relies on expensive Monte Carlo estimate to approximate the twist function. 
    }
    \label{fig:concept_figure}
\end{figure}

%% file: Algorithms/smc.tex
\SetKwInput{KwInput}{Input}
\SetKwInput{KwOutput}{Output}
\SetKwProg{Fn}{Function}{}{}

\begin{algorithm}[t]
    \SetAlgoLined
    \DontPrintSemicolon
    \setstretch{1.2}
    \caption{Twisted Sequential Monte Carlo / Importance Sampling}
    \label{alg:smc}
    \Fn{\textnormal{TwistSMC}($K$, $q$, $\Psi$, $\mathrm{ESS}_\mathrm{thres}$, $t_{\text{stop}}$)}{
        \tcp{Inputs:}
        \tcp{$K$: Number of particles \quad $q$: Proposal distribution}
        \tcp{$\Psi$: Twist function (MC estimate $\hat{\psi}$ or network $\psi^\phi$)}
        \tcp{At inference, set $\Psi_0(\B{x}_0)=\exp(r(\B{x}_0)/\beta)$.}
        \tcp{$\mathrm{ESS}_\mathrm{thres}$: Resampling threshold ($0$ for IS without resampling)}
        \tcp{$t_{\text{stop}}$: Stop timestep}
        
        \vspace{0.3em}
        $\{\B{x}_T^{(k)}\}_{k=1}^K \sim p_T^{\text{base}}, \quad \{W_{T}^{(k)}\}_{k=1}^K \leftarrow 1$ \;
        \For{$t = T, \dots, t_{\text{stop}} + 1$}{
            $\{\B{x}_{t-1}^{(k)}\}_{k=1}^K \sim \{q(\cdot \mid \B{x}_t^{(k)})\}_{k=1}^K$ \;
            
            $\{w_{t-1}^{(k)}\}_{k=1}^K \leftarrow \left\{ \dfrac{\Psi_{t-1}(\B{x}_{t-1}^{(k)})\, p^{\text{base}}(\B{x}_{t-1}^{(k)} \mid \B{x}_t^{(k)})}{\Psi_t(\B{x}_t^{(k)})\, q(\B{x}_{t-1}^{(k)} \mid \B{x}_t^{(k)})} \right\}_{k=1}^K$ \tcp*{\cref{eq:smc_weight} }
            
            $\{W_{t-1}^{(k)}\}_{k=1}^K \leftarrow \left\{ W_{t}^{(k)} \cdot w_{t-1}^{(k)} \right\}_{k=1}^K$ \;
            
            $\{\tilde{W}_{t-1}^{(k)}\}_{k=1}^K \leftarrow \left\{ W_{t-1}^{(k)} \Big/ \sum_{j=1}^{K} W_{t-1}^{(j)} \right\}_{k=1}^K$ \;
            
            $\mathrm{ESS} \leftarrow 1 \Big/ \sum_{k=1}^{K} (\tilde{W}_{t-1}^{(k)})^2$ \;
            
            \If{$\mathrm{ESS} < K\,\mathrm{ESS}_\mathrm{thres}$}{
                $\{a^{(k)}_{t-1}\}_{k=1}^K \sim \mathrm{Cat}\!\left(\{\tilde{W}_{t-1}^{(i)}\}_{i=1}^{K}\right)$ \; 
                $\{\B{x}_{t-1}^{(k)}\}_{k=1}^K \leftarrow \{\B{x}_{t-1}^{(a^{(k)}_{t-1})}\}_{k=1}^K, \quad \{W_{t-1}^{(k)}\}_{k=1}^K \leftarrow 1$ \tcp*{ Resampling }
            }
        }
        \Return $\{(\B{x}_{t_{\text{stop}}}^{(k)}, \tilde{W}_{t_{\text{stop}}}^{(k)})\}_{k=1}^{K}$ 
    }
\end{algorithm}

%% file: Algorithms/cdm.tex
\begin{algorithm}[t]
    \SetAlgoLined
    \DontPrintSemicolon
    \setstretch{1.2}
    \caption{\Ours{}: Contrastive Twist Training via Diffusion Forward Kernel}
    \label{alg:cdm_training}
    \KwInput{
        Base model $p^{\text{base}}$, MC twist $\hat{\psi}$~\cref{eq:mc_approx}, twist network $\psi^\phi$, ESS threshold $\mathrm{ESS}_\mathrm{thres}$, positive buffer size $B_{\text{buffer}}$, update interval $n_\text{update}$, learning rate $\gamma$, EMA rate $\eta$, batch size $B$, $\overline{\phi} = \texttt{stopgrad}(\phi)$
    }
    \KwOutput{Optimized twist network parameters $\phi$}
    $\phi_{\text{EMA}} \leftarrow \phi, \quad \text{step} \leftarrow 0, \quad \mathcal{B} \leftarrow \emptyset$ \;
    \While{not converged}{
        $\text{step} \leftarrow \text{step} + 1$ \;
        $t \sim \mathcal{U}(\{1, \dots, T\})$ \;
        
        \vspace{0.3em}
        \If{$\mathrm{step} \ \% \ n_\text{update} = 0$ \textnormal{\textbf{or}} $\mathcal{B} = \emptyset$}{
            $\{(\B{x}_{0}^{*(k)}, \tilde{W}_0^{*(k)})\}_{k=1}^{B_{\text{buffer}}} \leftarrow \text{TwistSMC}(B_{\text{buffer}}, p^{\text{base}}, \hat{\psi}, \mathrm{ESS}_\mathrm{thres}, 0)$ \tcp*{Pos. \cref{alg:smc} }
            $\mathcal{B} \leftarrow \{(\B{x}_0^{*(k)}, \tilde{W}_0^{*(k)})\}_{k=1}^{B_{\text{buffer}}}$ \;
        }
        $\{(\B{x}_{0}^{*(i)}, \tilde{W}_0^{*(i)})\}_{i=1}^B \sim \mathcal{B}$ \;
        $\{\B{x}_{t}^{*(i)}\}_{i=1}^B \sim \{p^{\text{base}}(\cdot \mid \B{x}_{0}^{*(i)})\}_{i=1}^B$ \tcp*{Forward~\cref{eq:forward}}
        
        \vspace{0.3em}
        $\{(\B{x}_{t}^{\overline{\phi}(i)}, \tilde{W}_t^{\phi(i)})\}_{i=1}^B \leftarrow \text{TwistSMC}(B, p^{\text{base}}, \psi^{\phi_{\text{EMA}}}, 0, t)$ \tcp*{Neg. \cref{alg:smc} }
        
        \vspace{0.3em}
        $\nabla_\phi \mathcal{L}_{\text{\Ours{}}} \leftarrow -\tfrac{1}{B} \sum_{i=1}^B \left( \tilde{W}_0^{*(i)} \nabla_\phi \log \psi_t^\phi(\B{x}_{t}^{*(i)}) - \tilde{W}_t^{\phi (i)}\nabla_\phi \log \psi_t^\phi(\B{x}_{t}^{\overline{\phi}(i)}) \right)$ \tcp*{\cref{eq:cdm_grad_decomp} }
        $\phi \leftarrow \phi - \gamma \, \nabla_\phi \mathcal{L}_{\text{\Ours{}}}$ \;
        $\phi_{\text{EMA}} \leftarrow \eta \, \phi_{\text{EMA}} + (1 - \eta) \, \phi$ \;
    }
    \Return $\phi$
\end{algorithm}

%% file: Sections/A3_Implementation_Details.tex
\section{Implementation Details}
\label{sec:impl_detail}

In this section, we introduce choice of twist network parameterization and present the implementation details of the baselines including \Ours{}.

\input{Figures/twist_head}
\input{Tables/runtime}

\subsection{Twist Head Parameterization}
\label{subsec:twist_parameterization}

As discussed in \cref{subsec:efficiency}, instead of training a separate network from scratch for the twist function, we parameterize the twist function as a lightweight head as shown in \cref{fig:twist_head}. 
The twist head takes as input the last hidden state of the denoising network and outputs a scalar representing the value.

To validate our claim, we report a runtime analysis in~\cref{tab:runtime_analysis}, where we use NVIDIA PRO $6000$ to measure the runtime. 
The twist head forward pass accounts for less than $5\%$ of the base model runtime, and as little as $0.5\%$ in some configurations. 
This efficiency follows from the parameterization itself, since a single forward pass through the denoising backbone produces a shared feature representation from which the logit head and twist head are computed in parallel. 
Consequently, evaluating $\psi_t^\phi$ during SMC adds negligible cost relative to standard sampling.

As shown in \cref{fig:twist_head}, we consider three different architectural choices for the twist head: MLP, MLP+PE, and Transformer.  

\textbf{MLP.} The MLP takes as input the mean-pooled representation of the last hidden state of the denoising network.

\textbf{MLP+PE.} We incorporate positional embeddings into the twist model. Specifically, we take the final hidden representations from the denoising network and add sinusoidal positional embeddings to each token. A shared MLP head is then applied token-wise to produce positionally encoded token features. The resulting representations are mean pooled across tokens and passed through a final MLP.

\textbf{Transformer.} For tasks with stronger positional dependencies, we also consider a transformer-based architecture~\cite{vaswani2017:attention} to more effectively capture the positional dependencies.

The choice of the twist head architecture depends on the information already encoded by the backbone. 
An MLP head suffices when the backbone features carry positional and contextual information (\eg~DiT and DPLM-2), whereas CNN backbones benefit from additional positional encodings (MLP+PE), and dLLMs from a lightweight Transformer head. 

For fair comparison, we use the same twist head architecture and shared training hyperparameters across \svdd{} and \Ours{} in the evaluations reported in~\cref{sec:experiments}.

\input{Tables/hyperparameters}

\subsection{Experiment Details}
\cref{tab:training_hparams} summarizes the hyperparameters used in each of the four experiments in \cref{sec:experiments}. 
For \Ours{}, we include the size of the positive buffer as well as the the positive buffer update interval $n_\text{update}$ which specifies the number of training steps taken before the positive buffer is cleared and resampled. 

For all experiments except the dLLM alignment task, we trained the twist estimator on a single NVIDIA RTX Pro 6000 GPU. For dLLM alignment, we used 4 NVIDIA B200 GPUs.

\paragraph{Hyperparameter Selection.}
The KL weight $\beta$ interacts with the numerical range of the reward, since the twist $\psi_t = \exp(V_t/\beta)$ becomes exponentially large for a small $\beta$, while a large $\beta$ over-regularizes toward the base model and weakens alignment. 
In our experiments, $\beta$ is set such that $\psi_t$ stays roughly within $[0, 10]$. 
The buffer update interval $n_\text{update}$ controls how many gradient steps each positive sample is reused for, where a larger $n_\text{update}$ reduces the number of reward evaluations during training and is thus beneficial when the reward is expensive. 
The Monte Carlo sample size $M$ trades off the accuracy of the twist estimate against compute. 
Lastly, the number of diffusion steps $T$ is inherited from the default setting of the base model.

\paragraph{Toxic Text Generation.}
For the toxic text generation task, we use a publicly available pretrained MDM~\cite{Sahoo:2024MDLM} trained on the OpenWebText dataset~\cite{Gokaslan2019:OpenWeb}. 
Following prior work~\cite{ou2025:smc_amot}, we set the KL weight to $\beta = 0.2$ and generate sequences of length $100$ over $100$ denoising steps. 
In this task, we use an MLP twist head with $5$ layers and width $768$.

\paragraph{Regulatory DNA Sequence Design.}
We train a CNN-parameterized MDM~\cite{Sahoo:2024MDLM} on a publicly available enhancer activity dataset~\cite{Gosai:2023MPRA}. 
For all experiments, we generate DNA sequences with a fixed sequence length of $200$, and run the diffusion process for $50$ steps during sampling. 
Since the denoising backbone is CNN-parameterized, the twist head incorporates positional information explicitly via an MLP with sinusoidal positional encodings (MLP+PE), comprising $7$ layers of width $1024$.

\paragraph{Protein Designability.}
We evaluate our method on protein generation using DPLM-2~\cite{Wang:2025DPLM}, a discrete diffusion model that jointly generates amino acid sequences and structural tokens. 
For all experiments, we generate proteins with a fixed sequence length of $100$ residues ($204$ tokens total, including $100$ sequence tokens, $100$ structure tokens, and $4$ \texttt{<bos>} and \texttt{<eos>} tokens for the sequence and structure tokens). 
The denoising step is set to $20$, which we found to provide a good balance between generation quality and computational efficiency. 
We use a larger buffer update interval $n_\mathrm{update} = 10$ in protein generation, since reward evaluation is expensive to amortize the cost of positive sampling over more gradient steps. 
We parameterize the twist head as a $5$-layer MLP with width $1024$. 
The twist model is trained using a learning rate of $1 \times 10^{-4}$ and weight decay $0.01$ using the AdamW optimizer~\cite{loshchilov2017:adamw}.

\paragraph{dLLM Alignment.}
For the dLLM alignment task, we adopt the pretrained LLaDA-8B-Instruct~\cite{nie:2025llada} as our base diffusion language model. We train the twist estimator using prompts from RewardBench~\cite{lambert:2025rewardbench}, randomly partitioned into an 80\% training split and a 20\% validation split for evaluation.
In this experiment, we generate sequences of length $128$ using $128$ denoising steps and set the KL regularization weight to $\beta=2.0$, a larger value than in the other applications because the reward model produces reward scores on a larger scale.
We parameterize the twist head as a $2$-layer Transformer~\cite{vaswani2017:attention} with hidden dimension $4096$.

%% file: Figures/twist_head.tex
\begin{figure}[t!]
    \centering
    \includegraphics[width=\linewidth]{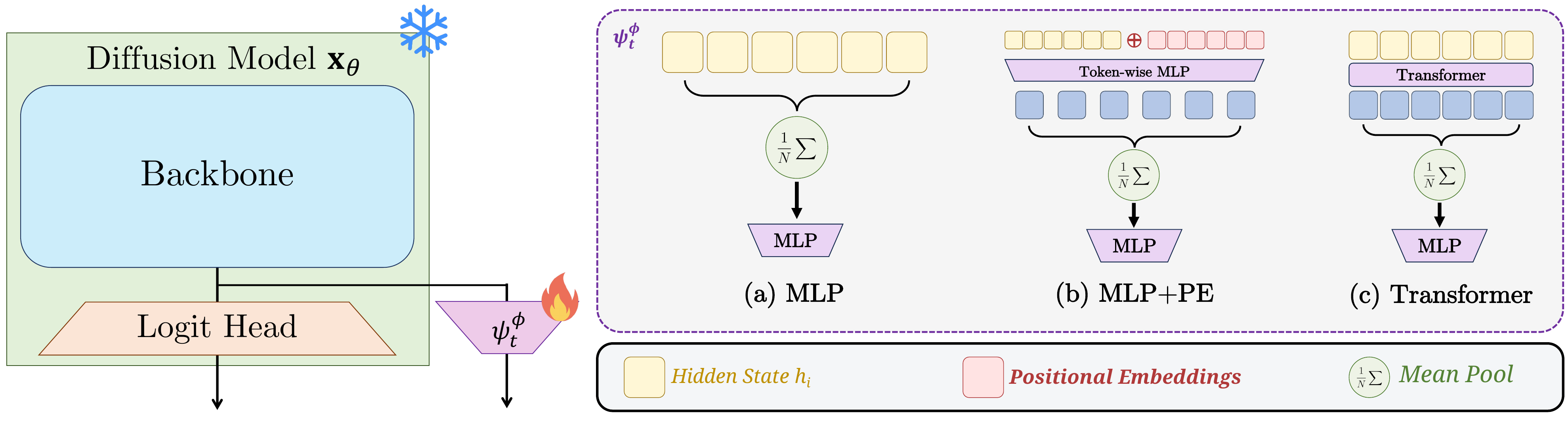}
    \caption{\textbf{Twist Head Architecture.} \textbf{(Left)} We parameterize the twisting function as a lightweight head that predicts the value based on the last hidden state of the denoising network. \textbf{(Right)} We consider three architectural choices for the twist head: (a) MLP, (b) MLP+PE, and (c) Transformer}
    \label{fig:twist_head}
\end{figure}

%% file: Tables/runtime.tex
\begin{table}[t!]
    \centering
    \small
    \setlength{\tabcolsep}{2.9pt}
    \renewcommand{\arraystretch}{1.2}

    \newcolumntype{Z}{>{\centering\arraybackslash}m{0.35\linewidth}}
    \newcolumntype{C}{>{\centering\arraybackslash}m{0.15\linewidth}}
    
    \begin{tabularx}{\linewidth}{Z | C C C C}
        \toprule
         & \makecell{\textbf{Base Model} \\ $\B{x}_\theta$ (ms)} & \makecell{\textbf{Twist Head} \\ $\psi^\phi$ (ms)} & \makecell{\textbf{Reward Model} \\ $r$ (ms)}  & \makecell{ \textbf{Twist / Base} \\ \textbf{Ratio (\%)}} \\
        \midrule
        \textbf{Toxic Text Generation} (\cref{subsec:toxic_text}) & 14.227  & 0.101  & 4.414  & 0.710 \\
        \textbf{Regulatory DNA Design} (\cref{subsec:dna_design})      & 9.829   & 0.289  & 11.438  & 2.940 \\
        \textbf{Protein Designability} (\cref{subsec:protein_design})  & 33.901  & 0.144  & 1356.0  & 0.425 \\
        \textbf{dLLM Alignment} (\cref{subsec:dllm_align})    & 28.619  & 1.327  & 24.268  & 4.637 \\
        \bottomrule
    \end{tabularx}
    \vspace{0.5\baselineskip}
    \caption{\textbf{Twist Function Runtime Analysis.} 
        We parameterize the twist function as a lightweight scalar head, incurring negligible runtime overhead. 
    }
    \label{tab:runtime_analysis}
\end{table}

%% file: Tables/hyperparameters.tex
\begin{table}[t]
    \centering
    \small
    \setlength{\tabcolsep}{2.9pt}
    \renewcommand{\arraystretch}{1.2}

    \newcolumntype{Z}{>{\centering\arraybackslash}m{0.35\linewidth}}
    \newcolumntype{C}{>{\centering\arraybackslash}m{0.15\linewidth}}
    
    \begin{tabularx}{\linewidth}{Z | C C C C}
        \toprule
         & \makecell{\textbf{Toxic Text} \\ \textbf{Generation}} & \makecell{\textbf{Regulatory} \\ \textbf{DNA design}} & \makecell{\textbf{Protein} \\ \textbf{Designability}} & \makecell{\textbf{dLLM} \\ \textbf{Alignment}} \\
        \midrule
        \textbf{Sequence Length ($N$)} & $100$ & $200$ & $204$ & $128$ \\
        \textbf{Diffusion Steps} ($T$) & $100$ & $50$ & $20$ & $128$ \\
        \textbf{Norm. ESS} ($\mathrm{ESS}_\mathrm{thres}$) & $0.5$ & $0.5$ & $0.5$ & $0.5$ \\
        \textbf{KL Weight} ($\beta$) & $0.2$ & $0.2$ & $0.1$ & $2.0$ \\
        \midrule
        \textbf{Architecture} & MLP & MLP + PE & MLP & Transformer \\
        \midrule
        \textbf{Optimizer} & AdamW & AdamW & AdamW & AdamW \\
        \textbf{Learning Rate} ($\gamma$) & $1\mathrm{e}{-4}$ & $3\mathrm{e}{-4}$ & $1\mathrm{e}{-4}$ & $1\mathrm{e}{-4}$ \\
        \textbf{Weight Decay} & $0.01$ & $0.01$ & $0.01$ & $0.01$ \\
        \midrule
        \makecell{\textbf{Buffer Size} ($B_{\text{buffer}}$)} & $64$ & $64$ & $64$ & $64$ \\
        \makecell{\textbf{Buffer} \textbf{Update Interval} ($n_\text{update}$)} & $4$ & $4$ & $10$ & $4$ \\
        \bottomrule
    \end{tabularx}
    \vspace{0.5\baselineskip}
    \caption{
    \textbf{Training and Inference Hyperparameters.} 
    We fix the hyperparameters across all baselines to ensure a fair comparison. 
    \svdd{} and \Ours{} use the same twist function parameterization, while \Ours{} introduces additional hyperparameters for the positive buffer. 
    }
    \label{tab:training_hparams}
\end{table}

%% file: Sections/A4_Additional_Results.tex
\section{Ablation Studies}
\label{sec:ablation}

In this section, we provide ablation and hyperparameter studies for \Ours{} training.

\input{Figures/ema_pos_buff_ablation}
\input{Tables/ablation}

\paragraph{Positive Buffer.}
As discussed in \cref{sec:experiments}, for more efficient training, we reuse the positive samples by maintaining a buffer $\mathcal{B}$ and re-noising them across timesteps. 
The positive buffer is updated every $n_\text{update}$ steps. 
Increasing $n_\text{update}$ improves training efficiency by reusing positive samples obtained from SMC. 
In particular, when the reward is expensive, increasing $n_\text{update}$ reduces the number of reward evaluations required. 
In \cref{fig:toxicity-nupdate,fig:dna-nupdate}, we show that \Ours{} performs well across various update intervals $n_\text{update}$.

\paragraph{Monte Carlo Sample Size.}
Although twist-learning methods like \svdd{} and \Ours{} bypass the need for Monte Carlo estimation at inference by amortizing the cost into a single forward pass, the number of Monte Carlo samples $M$ remains as a hyperparameter during training. \Cref{tab:mc_sample} presents the quantitative performance of both methods across varying values of $M$ used for target estimation at train time. 
For \svdd{}, increasing $M$ initially yields performance gains. 
However, as shown in \cref{fig:train_convergence}, these improvements quickly plateau for $M > 1$. 
In contrast, \Ours{} maintains consistent performance across different values of $M$.

\paragraph{Choice of Sampling.}
As discussed in \cref{subsec:cdm}, positive sampling during the training of \Ours{} can be performed using either standard Importance Sampling (IS) or Sequential Monte Carlo (SMC). 
As shown in \cref{tab:sampling_choice}, using SMC for positive sampling consistently yields better downstream performance than IS under identical hyperparameter settings. 
We attribute this improvement to the resampling mechanism of SMC, which mitigates the severe weight degeneracy often observed in IS. 

\paragraph{Number of Positive Samples.}
We analyze the robustness of \Ours{} to the finite number of positive samples used during training by varying the total number of distinct positive samples to $6.25\%$, $25\%$, $50\%$, and $100\%$ of the default setup. 
As shown in~\cref{tab:pos_sample}, \Ours{} shows relatively robust performance with $50\%$ of the positive samples on both toxic text generation and regulatory DNA design. 
Beyond this point, the learned twist becomes less efficient as the number of positive samples decreases, requiring a larger number of particles to achieve the similar level of quality. 
In the extreme case of using only $6.25\%$ of the positive samples, both tasks show noticeable degradation, which is not recovered even with a larger number of particles. 

\paragraph{Reward Range of Positive Samples.}
To examine how the learned twist generalizes beyond the positive samples in the buffer, we restrict the positive samples to a narrow reward range, using only the samples that score in the bottom $10\%$ (Low), the middle $10\%$ (Mid), and the top $10\%$ (Top) of the reward. 
We additionally report the case using the full positive buffer (Full) as a reference. 
As shown in~\cref{tab:pos_range}, training \Ours{} with a narrow band of rewards may bias the learned twist and degrade performance. 
While using the top $10\%$ samples yields relatively robust performance, using low-reward samples results in noticeable drops. 
In practice, the full positive buffer covers a sufficient range of rewards, such that the learned twist remains robust at inference. 

\input{Tables/positive_buffer}

\paragraph{Twist-Guided Positive Sampling.}
To further reduce the cost of positive sampling, we consider replacing the reward queries at the intermediate steps of positive sampling with the learned twist function, while reweighting with the given reward at the terminal step for correction. 
We compare the following cases: (1) twist-guided SMC, where the learned twist replaces the Monte Carlo reward estimate at every intermediate step of positive sampling, (2) Reward $\rightarrow$ Twist ($25\%$, $50\%$, and $75\%$), where the Monte Carlo reward estimate is used for the given fraction of the training iterations, after which positive sampling switches to twist-guided SMC, and (3) \Ours{}, which uses the Monte Carlo reward estimate throughout training and corresponds to case (2) at $100\%$. 
As shown in~\cref{tab:twist_pos}, twist-guided SMC underperforms, which we attribute to the twist function being severely undertrained in early iterations, such that the resulting positive samples in the buffer are largely misaligned with the target. 
On the other hand, we observe consistent improvements as the fraction of the Monte Carlo reward estimate increases ($25\% \rightarrow 50\% \rightarrow 75\%$), which trades off training compute against the downstream reward. 

\input{Tables/twist_positive_sampling}

\paragraph{Sampling Distribution for \svdd{}.}
To isolate the contribution of the contrastive objective, we ablate the sampling process used in the regression objective of \svdd{} in~\cref{eq:regression}. 
We consider the following cases: (1) base-sample regression ($\B{x}_t \sim p_t^\text{base}$), which is the default \svdd{} baseline trained on samples from the base distribution, (2) target-sample regression ($\B{x}_t \sim p_t^*$), which draws $\B{x}_t$ from the approximate target and regresses $\psi_t^\phi(\B{x}_t)$ onto the Monte Carlo twist estimate, (3) forward-noised target regression ($\B{x}_t \sim p^\text{base}(\B{x}_t \mid \B{x}_0^*)$), which applies the forward kernel to case (2) to amortize training as in~\cref{eq:cdm_grad_decomp}, and (4) \Ours{}, which uses the contrastive objective with forward-kernel amortization. 
Note that case (2) requires a separate SMC run to draw $\B{x}_t$ at every timestep, which is less efficient than case (3) that utilizes our forward-based amortization. 
As shown in~\cref{tab:sv_ablation}, case (2) fails to generate high-reward samples, likely due to its inefficient sampling process. 
For toxic text generation, case (3) outperforms case (1) but still falls short of \Ours{}, whereas for regulatory DNA design, case (3) performs comparably to or slightly worse than case (1). 

\input{Tables/soft_value_ablation}

\section{Additional Results}
\label{sec:additional_results}

In this section, we provide additional results on fine-tuned proposals, training convergence, particle sizes, effective sample size, and reward sensitivity.

\paragraph{Additional Results on Compatibility with Fine-Tuned Proposal.}

\begin{figure}[t!]
\centering

\begin{minipage}{0.48\textwidth}
    \centering
    \includegraphics[width=\textwidth]{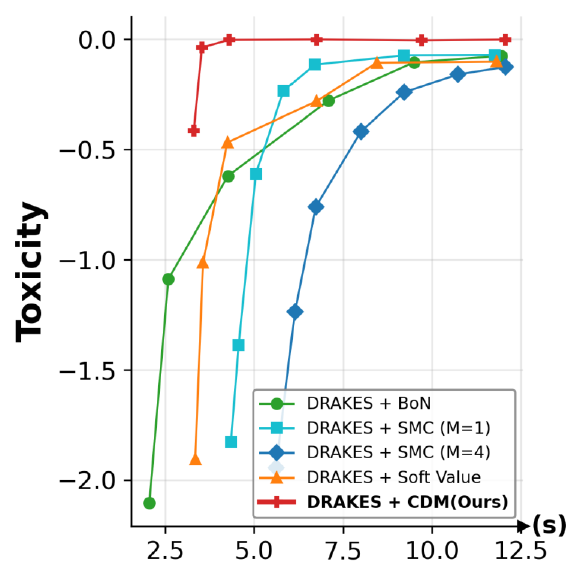} 
    \captionof{figure}{
    \textbf{Scaling with Direct Backpropagation Fine-Tuned Proposal~\cite{wang2024:drakes}.}
    \Ours{} is also compatible with DRAKES~\cite{wang2024:drakes}, a proposal fine-tuned via direct backpropagation for tasks with differentiable rewards. 
    }
    \label{fig:drakes_toxicity}
\end{minipage}
\hfill
\begin{minipage}{0.48\textwidth}
    \centering
    \includegraphics[width=\textwidth]{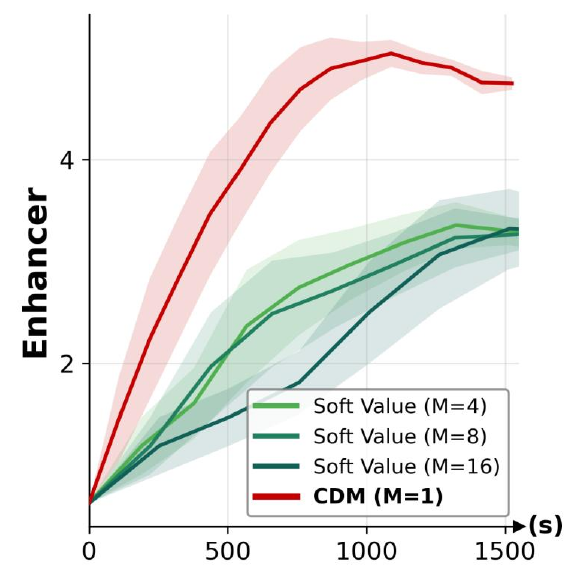}
    \captionof{figure}{
    \textbf{Training Comparison of \Ours{} and \svdd{} for Regulatory DNA Sequence Design.}
    \Ours{} shows faster training convergence and achieves higher reward than \svdd{} with different $M$. 
    }
    \label{fig:dna_train_dynamics}
\end{minipage}

\end{figure}

For tasks with differentiable rewards, such as toxic text generation (\cref{subsec:toxic_text}), reward alignment can be achieved by fine-tuning the proposal via direct gradient backpropagation. 
To explore this regime, we additionally pair \Ours{} with DRAKES~\cite{wang2024:drakes}, a method that fine-tunes a base model through direct backpropagation using the Gumbel-Softmax trick~\cite{jang:2016categorical}. 
As shown in \cref{fig:drakes_toxicity}, applying \Ours{} on top of DRAKES further improves its performance, outperforming all other baselines. 
This demonstrates the compatibility and synergistic potential of \Ours{} when integrated with fine-tuned models.

\paragraph{Additional Results on Training Convergence.}
In addition to the quantitative results discussed in \cref{subsec:twist_exp}, we present comparison of the training dynamics of \svdd{} and \Ours{} for the regulatory DNA sequence design task in \cref{fig:dna_train_dynamics}. 
We observe a consistent trend: \svdd{} plateaus early in training, whereas \Ours{} converges more efficiently and achieves a higher final reward. 

\paragraph{Effect of Training and Inference Particle Sizes.}
Following the analysis in Appendix~\ref{subsec:finite_particle}, we report the performance of \Ours{} on toxic text generation with varying numbers of particles used for positive sampling during training ($K_\text{train}$) and at inference ($K_\text{inf}$) in~\cref{tab:particle_size}. 
Consistent with our analysis, the performance of \Ours{} improves as both $K_\text{train}$ and $K_\text{inf}$ increase. 

\input{Tables/particle_size}

\paragraph{Effective Sample Size (ESS).}
We report the mean normalized ESS $\in (0, 1]$ along the denoising schedule and the resampling rate of \Ours{} for each application in~\cref{tab:ess}, where resampling is performed when the normalized ESS falls below $0.5$ (\cref{tab:training_hparams}). 
We observe that the ESS remains low for some applications. 
While the high dimensionality of the state space may play a role in lowering the ESS, we also observe that the ESS can be large when the proposal distribution is well aligned with the target distribution. 
For instance, the protein designability task shows a high ESS, which is also indicated by the relatively favorable scaling of BoN (\cref{subsec:protein_design}). 
To this end, the learned twist can be combined with a fine-tuned proposal for synergistic gains, as shown in~\cref{subsec:finetune_exp}. 

\input{Tables/ess}

\paragraph{Reward Sensitivity.}
When the reward is nearly constant over the support of the base model, \ie~$p_t^* \approx p_t^\text{base}$, the positive and negative distributions in~\cref{eq:cdm_grad} coincide and the gradient vanishes. 
This is the expected behavior, since there is no signal to align with. 
When the high-reward region is sparse, the SMC sampler can exhibit high variance and the resulting positive samples are poorly aligned with the target, in which case \Ours{} can struggle to align with the target, as do most inference-time scaling and post-training methods. 
To indicate the regime in which \Ours{} is evaluated, we report the reward statistics of the base model samples for each application in~\cref{tab:reward_stats}. 

We further evaluate \Ours{} with different toxicity classifiers as the given reward, including the Detoxify~\cite{hanu:2020detoxify} variants (Original, Unbiased, and Multilingual) and the default Paradetox classifier~\cite{logacheva:2022paradetox}. 
As shown in~\cref{tab:reward_models}, \Ours{} shows consistent scaling with increasing particle size regardless of the reward model. 
Since the reported accuracies of these classifiers are evaluated on different benchmarks, we additionally simulate an imperfect reward model by perturbing the outputs of the default classifier with Gaussian noise (Paradetox + Noise). 
As expected, the performance degrades with the perturbed reward, indicating that \Ours{} can suffer when the reward model is misaligned. 

\input{Tables/reward_sensitivity}

\section{Amortization Cost Analysis}
\label{sec:cost_analysis}

In this section, we analyze the regime in which amortization pays off. 
We first compare the per-sample cost of \Ours{} against vanilla SMC inference-time scaling. 
We then account for the upfront training cost and derive the break-even number of downstream samples $n^*$ beyond which amortization is cheaper. 
Finally, we instantiate the analysis for each application. 
Note that both \Ours{} and inference-time scaling draw samples from the same backbone and differ only in how the twist is obtained. 
Hence, the training cost of the backbone is common to both and does not enter this comparison. 

\paragraph{Notation and Assumptions.}
Let $T$ be the number of denoising steps, $K$ the number of particles, and $M$ the number of Monte Carlo samples per twist estimate in~\cref{eq:mc_approx}. 
We denote by $c_\theta$, $c_\psi$, and $c_r$ the cost of one backbone forward pass, one twist-head forward pass, and one reward evaluation, respectively, and by $n_\text{pos}$ and $n_\text{neg}$ the total number of positive and negative samples consumed during training. 
We make the following assumptions: 
\begin{itemize}[leftmargin=*, itemsep=2pt, topsep=2pt]
    \item The same particle size $K$ is used for \Ours{} training and for both \Ours{} and SMC inference. This is a mild assumption, as one typically targets the quality of $K$-particle SMC and sets the training particle size of \Ours{} accordingly. 
    \item The cost of re-noising a buffer sample with the forward kernel is negligible. 
    \item The cost of the backward pass of the twist head is negligible compared to the negative sampling cost. 
\end{itemize}

\paragraph{Per-Sample Inference Cost.}
We report the cost per generated sample and omit $K$ for simplicity, as $K$ particles cost $K$ times as much and return $K$ weighted samples. 
Generating one sample requires $T$ backbone forward passes. 
On top of this, SMC estimates the twist at every step with $M$ Monte Carlo samples (\cref{eq:mc_approx}), each requiring a reward evaluation:
\begin{equation}
    c_\text{SMC} = T c_\theta + T M c_r. 
\end{equation}
Note that this is also exactly the cost of drawing one positive sample during \Ours{} training, since positive samples are drawn from the same Monte Carlo twist SMC procedure. 
At inference, \Ours{} replaces the per-step Monte Carlo estimate with a single twist-head evaluation, and calls the reward only once at the terminal step $t=0$, where $\psi_0^*(\B{x}_0) = \exp(r(\B{x}_0)/\beta)$ corrects the final weight:
\begin{equation}
    c_\text{\Ours{}} = T c_\theta + T c_\psi + c_r. 
\end{equation}
The per-sample gain is therefore given as:
\begin{equation}
    \Delta \coloneqq c_\text{SMC} - c_\text{\Ours{}} = (TM - 1) c_r - T c_\psi. 
\end{equation}
\Ours{} is cheaper per sample whenever $\Delta > 0$, \ie~whenever $c_r / c_\psi > T / (TM - 1) \approx 1/M$. 
As expected, the gain grows with the reward cost $c_r$ and shrinks with the twist-head cost $c_\psi$. 
Since the twist head costs a small fraction of a single backbone forward pass (\cref{tab:runtime_analysis}), this condition holds in all of our applications. 

\paragraph{Upfront Training Cost and Break-Even Sample Count.}
Positive sampling is the only part of \Ours{} training that calls the reward, at cost $c_\text{SMC}$ per positive sample. 
Negative samples are drawn from the base process tilted by the learned twist down to a timestep $t$ drawn uniformly from $\{1, \dots, T\}$, and hence traverse $T/2$ steps in expectation:
\begin{equation}
    c_\text{neg} = \frac{T}{2} \left( c_\theta + c_\psi \right). 
\end{equation}
The total upfront cost of \Ours{} is $C_\text{train} = n_\text{pos} c_\text{SMC} + n_\text{neg} c_\text{neg}$, and amortization pays off once the number of downstream samples exceeds:
\begin{equation}
    n^* = \frac{C_\text{train}}{\Delta} = \frac{n_\text{pos} c_\text{SMC} + n_\text{neg} c_\text{neg}}{(TM - 1) c_r - T c_\psi}. 
\end{equation}
At a fixed number of gradient updates, and hence a fixed $n_\text{neg}$, a larger $n_\text{update}$ requires fewer positive samples and therefore lowers both $C_\text{train}$ and $n^*$. 
We consider the following two cases. 
When the reward dominates ($c_r \gg c_\theta$), we have $c_\text{SMC} \rightarrow TMc_r$ and $\Delta \rightarrow TMc_r$, while $c_\text{neg}$ becomes negligible, so that $n^* \rightarrow n_\text{pos}$. 
\Ours{} then pays for itself after roughly as many downstream samples as the positive samples it consumed during training. 
Conversely, when the reward is cheap ($c_r \lesssim c_\theta$), the reward-free negative sampling cost no longer vanishes and $\Delta$ shrinks, so that $n^*$ grows to a multiple of $n_\text{pos}$.

We instantiate this analysis in each application presented in~\cref{sec:experiments} and report the break-even number of samples in~\cref{tab:cost_analysis}. 
The break-even point is governed by how expensive the reward is relative to a single backbone forward pass. 
When the reward dominates the cost, as in protein design, the upfront training cost is recovered after a relatively small number of downstream samples. 
This is also useful in practice, as the protein design process typically requires a large number of candidates before screening for wet-lab validation. 
On the other hand, when the reward is cheap and the backbone is relatively expensive, the per-sample saving is smaller and more samples are needed before training pays off. 
In these cases, running inference-time scaling directly can be preferable, particularly when only a small number of downstream samples is needed. 

\input{Tables/cost_analysis}

\section{Qualitative Results}
\label{sec:qual_result}

In this section, we provide qualitative results for protein and text generation.

\input{Figures/protein_main_qualitatives}

\subsection{Protein Designability}
\cref{fig:protein-qualitatives} shows the qualitative results for protein design. The generated protein is shown in blue while the refolded protein obtained using ESMFold~\cite{Lin:2023ESM} is shown in orange. We aim to generate designable proteins for which the generated and refolded structures are well aligned. \Ours{} achieves highly designable proteins, as shown by the overlapping and closely matching generated and refolded structures.

\subsection{Diffusion Large Language Model Alignment}
\Cref{tab:llada_qual_start}-\ref{tab:llada_qual_last} presents the qualitative results for diffusion LLM preference alignment~\cite{nie:2025llada}. 
We observe that \Ours{} generates responses that are highly aligned with user intents across diverse domains, including grammar, coding, and reasoning.

\begin{center}
    \hfill \break
    \textbf{Qualitative results are presented in the following pages.}
\end{center}

\input{Tables/llada_alignment}

%% file: Figures/ema_pos_buff_ablation.tex
\begin{figure}[t!]
    \centering
    \begin{subfigure}[t]{0.48\textwidth}
        \centering
        \includegraphics[width=\linewidth]{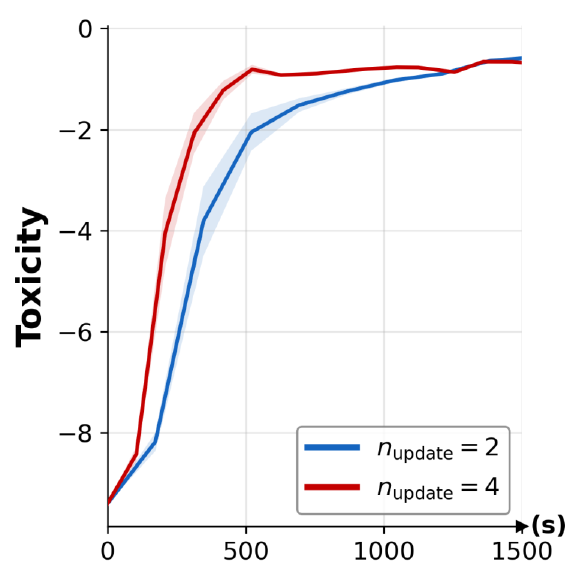}
        \caption{Toxic Text Generation with Varying $n_\text{update}$}\label{fig:toxicity-nupdate}
    \end{subfigure}\hfill
    \begin{subfigure}[t]{0.48\textwidth}
        \centering
        \includegraphics[width=\linewidth]{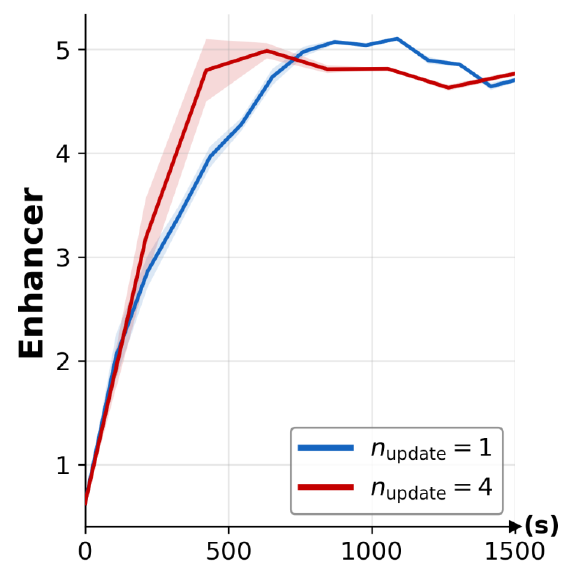}
        \caption{DNA Sequence Design with Varying $n_\text{update}$}\label{fig:dna-nupdate}
    \end{subfigure}\hfill
    \caption{
    \textbf{Positive Buffer Ablation Results.} 
    We present an ablation study on the buffer update frequency, $n_\mathrm{update}$, evaluating its impact on both (a) toxic text generation and (b) regulatory DNA sequence design. 
    }
    \label{fig:ablation}
\end{figure}

%% file: Tables/ablation.tex
\begin{table}[t!]
    \centering
    \setlength{\tabcolsep}{3pt}
    \renewcommand{\arraystretch}{1.1}

    \begin{subtable}[b]{0.65\linewidth}
        \centering
        \begin{tabularx}{0.49\textwidth}{l YY}
            \toprule
            \textbf{Toxicity} & $M=1$ & $M=4$ \\
            \midrule
            \svdd{} & -1.8060 & -0.9030 \\
            \Ours{}  & -0.4614 & -0.5079 \\
            \bottomrule
        \end{tabularx}
        \hfill %
        \begin{tabularx}{0.49\textwidth}{l YY}
            \toprule
            \textbf{Protein} & $M=1$ & $M=4$ \\
            \midrule
            \svdd{} & -1.8237 & -1.6895 \\
            \Ours{}  & -1.6950 & -1.5725 \\
            \bottomrule
        \end{tabularx}
        \caption{Monte Carlo Sample Size $M$ (Train-Time)}
        \label{tab:mc_sample}
    \end{subtable}
    \hfill
    \begin{subtable}[b]{0.33\linewidth}
        \centering
        \begin{tabularx}{0.9\textwidth}{l YY}
            \toprule
             & \textbf{Toxicity} & \textbf{Protein} \\
            \midrule
            IS  & -1.0367 & -1.8049 \\
            SMC & -0.4614 & -1.5725 \\
            \bottomrule
        \end{tabularx}
        \caption{Positive Sampling Method}
        \label{tab:sampling_choice}
    \end{subtable}

    \caption{\textbf{Choice of Monte-Carlo Sample Size $M$ and Positive Sampling Method.} (a) We show that \Ours{} outperforms Soft Value across various $M$. (b) While both IS and SMC can be used for sampling positive samples, we find that SMC yields better performance.}
    \label{tab:ablations}
\end{table}

%% file: Tables/positive_buffer.tex
\begin{table}[h!]
    \centering
    \small
    \setlength{\tabcolsep}{3pt}
    \renewcommand{\arraystretch}{1.1}

    \begin{subtable}[b]{\linewidth}
        \centering
        \begin{tabularx}{0.9\linewidth}{l c YYY YYY}
            \toprule
            & & \multicolumn{3}{c}{\textbf{Given} $\uparrow$} & \multicolumn{3}{c}{\textbf{Heldout} $\uparrow$} \\
            \cmidrule(lr){3-5} \cmidrule(lr){6-8}
            \textbf{Task} & \textbf{Positive Samples (\%)} & $K=4$ & $K=8$ & $K=16$ & $K=4$ & $K=8$ & $K=16$ \\
            \midrule
            \multirow{4}{*}{\makecell[l]{\textbf{Toxic Text}\\\textbf{Generation}}} 
            & $6.25$ & -4.270 & -2.434 & -1.576 & 0.497 & 0.703 & 0.797 \\
            & $25$   & -3.629 & -1.872 & -0.865 & 0.560 & 0.757 & 0.863 \\
            & $50$   & -2.284 & -1.109 & -0.387 & 0.747 & 0.837 & 0.930 \\
            & $100$  & -1.507 & -0.634 & -0.229 & 0.807 & 0.887 & 0.933 \\
            \midrule
            \multirow{4}{*}{\makecell[l]{\textbf{Regulatory}\\\textbf{DNA Design}}} 
            & $6.25$ & 1.756 & 1.737 & 1.645 & 1.693 & 1.605 & 1.577 \\
            & $25$   & 3.554 & 4.558 & 5.265 & 3.510 & 4.539 & 5.311 \\
            & $50$   & 4.025 & 5.063 & 5.616 & 4.017 & 5.119 & 5.824 \\
            & $100$  & 4.251 & 5.073 & 5.564 & 4.203 & 5.100 & 5.810 \\
            \bottomrule
        \end{tabularx}
        \caption{Number of Positive Samples}
        \label{tab:pos_sample}
    \end{subtable}

    \vspace{0.75\baselineskip}

    \begin{subtable}[b]{\linewidth}
        \centering
        \begin{tabularx}{0.9\linewidth}{l c YYY YYY}
            \toprule
            & & \multicolumn{3}{c}{\textbf{Given} $\uparrow$} & \multicolumn{3}{c}{\textbf{Heldout} $\uparrow$} \\
            \cmidrule(lr){3-5} \cmidrule(lr){6-8}
            \textbf{Task} & \textbf{Reward Range} & $K=4$ & $K=8$ & $K=16$ & $K=4$ & $K=8$ & $K=16$ \\
            \midrule
            \multirow{4}{*}{\makecell[l]{\textbf{Toxic Text}\\\textbf{Generation}}} 
            & Low  & -6.600 & -6.280 & -5.862 & 0.123 & 0.110 & 0.123 \\
            & Mid  & -2.135 & -1.305 & -1.185 & 0.673 & 0.707 & 0.707 \\
            & Top  & -1.625 & -0.651 & -0.239 & 0.843 & 0.890 & 0.910 \\
            & Full & -1.507 & -0.634 & -0.229 & 0.807 & 0.887 & 0.933 \\
            \midrule
            \multirow{4}{*}{\makecell[l]{\textbf{Regulatory}\\\textbf{DNA Design}}} 
            & Low  & 1.274 & 1.428 & 1.507 & 1.209 & 1.342 & 1.443 \\
            & Mid  & 3.565 & 3.786 & 3.831 & 3.599 & 3.939 & 4.053 \\
            & Top  & 4.072 & 4.786 & 4.904 & 4.053 & 4.712 & 4.965 \\
            & Full & 4.251 & 5.073 & 5.564 & 4.203 & 5.100 & 5.810 \\
            \bottomrule
        \end{tabularx}
        \caption{Reward Range of Positive Samples}
        \label{tab:pos_range}
    \end{subtable}

    \caption{\textbf{Positive Buffer Ablations.} 
    (a) \Ours{} trained with varying numbers of distinct positive samples, relative to the default setup. 
    (b) \Ours{} trained with positive samples restricted to the bottom (Low), middle (Mid), and top (Top) $10\%$ of the reward, compared with the full positive buffer (Full). 
    }
    \label{tab:positive_buffer}
\end{table}

%% file: Tables/twist_positive_sampling.tex
\begin{table}[h!]
    \centering
    \small
    \setlength{\tabcolsep}{3pt}
    \renewcommand{\arraystretch}{1.1}
    \begin{tabularx}{\linewidth}{l l YYY YYY}
        \toprule
            & & \multicolumn{3}{c}{\textbf{Given} $\uparrow$} & \multicolumn{3}{c}{\textbf{Heldout} $\uparrow$} \\
            \cmidrule(lr){3-5} \cmidrule(lr){6-8}
            \textbf{Task} & \textbf{Positive Sampler} & $K=4$ & $K=8$ & $K=16$ & $K=4$ & $K=8$ & $K=16$ \\
            \midrule
            \multirow{5}{*}{\makecell[l]{\textbf{Toxic Text}\\\textbf{Generation}}}
            & Twist-Guided SMC                & -7.180 & -6.390 & -5.429 & 0.214 & 0.277 & 0.366 \\
            & Reward $\rightarrow$ Twist ($25\%$) & -3.733 & -2.871 & -2.196 & 0.501 & 0.567 & 0.636 \\
            & Reward $\rightarrow$ Twist ($50\%$) & -2.671 & -1.124 & -0.560 & 0.687 & 0.876 & 0.931 \\
            & Reward $\rightarrow$ Twist ($75\%$) & -1.995 & -0.926 & -0.375 & 0.764 & 0.879 & 0.938 \\
            & \Ours{}                         & -1.507 & -0.634 & -0.229 & 0.807 & 0.887 & 0.933 \\
            \midrule
            \multirow{5}{*}{\makecell[l]{\textbf{Regulatory}\\\textbf{DNA Design}}}
            & Twist-Guided SMC                & 3.594 & 3.895 & 4.071 & 3.580 & 3.977 & 4.304 \\
            & Reward $\rightarrow$ Twist ($25\%$) & 4.068 & 4.698 & 4.733 & 4.109 & 4.720 & 5.116 \\
            & Reward $\rightarrow$ Twist ($50\%$) & 4.233 & 4.684 & 4.916 & 4.279 & 4.921 & 5.207 \\
            & Reward $\rightarrow$ Twist ($75\%$) & 4.264 & 4.807 & 5.237 & 4.335 & 4.958 & 5.496 \\
            & \Ours{}                         & 4.251 & 5.073 & 5.564 & 4.203 & 5.100 & 5.810 \\
            \bottomrule
    \end{tabularx}
    \vspace{0.5\baselineskip}
    \caption{\textbf{Twist-Guided Positive Sampling.} 
    We replace the Monte Carlo reward estimate in positive sampling with the learned twist throughout training (Twist-Guided SMC), or after a given fraction of the training iterations (Reward $\rightarrow$ Twist). 
    }
    \label{tab:twist_pos}
\end{table}

%% file: Tables/soft_value_ablation.tex
\begin{table}[h!]
    \centering
    \small
    \setlength{\tabcolsep}{3pt}
    \renewcommand{\arraystretch}{1.1}
    \begin{tabularx}{\linewidth}{l l YYY YYY}
        \toprule
            & & \multicolumn{3}{c}{\textbf{Given} $\uparrow$} & \multicolumn{3}{c}{\textbf{Heldout} $\uparrow$} \\
            \cmidrule(lr){3-5} \cmidrule(lr){6-8}
            \textbf{Task} & \textbf{Objective} & $K=4$ & $K=8$ & $K=16$ & $K=4$ & $K=8$ & $K=16$ \\
            \midrule
            \multirow{4}{*}{\makecell[l]{\textbf{Toxic Text}\\\textbf{Generation}}}
            & (1) \svdd{} (Base $\B{x}_t$) & -4.237 & -1.523 & -0.485 & 0.537 & 0.817 & 0.933 \\
            & (2) + Target $\B{x}_t$       & -7.678 & -6.252 & -4.689 & 0.167 & 0.270 & 0.413 \\
            & (3) + Forward Kernel         & -3.858 & -1.481 & -0.393 & 0.610 & 0.797 & 0.903 \\
            & (4) \Ours{}                  & -1.507 & -0.634 & -0.229 & 0.807 & 0.887 & 0.933 \\
            \midrule
            \multirow{4}{*}{\makecell[l]{\textbf{Regulatory}\\\textbf{DNA Design}}}
            & (1) \svdd{} (Base $\B{x}_t$) & 2.573 & 3.851 & 4.722 & 2.467 & 3.794 & 4.537 \\
            & (2) + Target $\B{x}_t$       & 1.404 & 1.716 & 2.093 & 1.324 & 1.574 & 1.939 \\
            & (3) + Forward Kernel         & 2.259 & 3.701 & 4.567 & 2.143 & 3.559 & 4.492 \\
            & (4) \Ours{}                  & 4.251 & 5.073 & 5.564 & 4.203 & 5.100 & 5.810 \\
            \bottomrule
    \end{tabularx}
    \vspace{0.5\baselineskip}
    \caption{\textbf{Sampling Distribution for \svdd{}.} 
    We train the regression objective of \svdd{} with samples drawn from the base distribution (1), the approximate target (2), and the forward-noised approximate target (3), and compare with \Ours{} (4). 
    }
    \label{tab:sv_ablation}
\end{table}

%% file: Tables/particle_size.tex
\begin{table}[h!]
    \centering
    \small
    \setlength{\tabcolsep}{3pt}
    \renewcommand{\arraystretch}{1.1}
    \begin{tabularx}{0.7\linewidth}{c YYY YYY}
        \toprule
        & \multicolumn{3}{c}{\textbf{Given} $\uparrow$} & \multicolumn{3}{c}{\textbf{Heldout} $\uparrow$} \\
        \cmidrule(lr){2-4} \cmidrule(lr){5-7}
        $K_\text{train}$ / $K_\text{inf}$ & $4$ & $8$ & $16$ & $4$ & $8$ & $16$ \\
        \midrule
        $4$  & -2.944 & -1.917 & -1.736 & 0.553 & 0.597 & 0.630 \\
        $8$  & -1.891 & -1.149 & -0.818 & 0.697 & 0.727 & 0.797 \\
        $16$ & -1.652 & -0.932 & -0.675 & 0.715 & 0.807 & 0.835 \\
        \bottomrule
    \end{tabularx}
    \vspace{0.5\baselineskip}
    \caption{\textbf{Effect of Training and Inference Particle Sizes.} 
    We report the given and heldout rewards of \Ours{} on toxic text generation with varying numbers of particles for positive sampling during training ($K_\text{train}$) and at inference ($K_\text{inf}$). 
    }
    \label{tab:particle_size}
\end{table}

%% file: Tables/ess.tex
\begin{table}[h!]
    \centering
    \small
    \setlength{\tabcolsep}{3pt}
    \renewcommand{\arraystretch}{1.1}
    \begin{tabularx}{\linewidth}{l YYYYYYYYYY c}
        \toprule
        & \multicolumn{10}{c}{\textbf{Denoising Progress} (\% of steps)} & \\
        \cmidrule(lr){2-11}
        & $10$ & $20$ & $30$ & $40$ & $50$ & $60$ & $70$ & $80$ & $90$ & $100$ & Resample Rate \\
        \midrule
        \textbf{Toxic Text Generation} & 0.12 & 0.18 & 0.22 & 0.24 & 0.25 & 0.24 & 0.18 & 0.14 & 0.09 & 0.07 & 0.92 \\
        \textbf{Regulatory DNA Design}        & 0.10 & 0.17 & 0.23 & 0.29 & 0.32 & 0.36 & 0.38 & 0.43 & 0.45 & 0.47 & 0.83 \\
        \textbf{Protein Designability}    & 0.36 & 0.77 & 0.77 & 0.78 & 0.81 & 0.81 & 0.81 & 0.81 & 0.81 & 0.81 & 0.07 \\
        \textbf{dLLM Alignment}       & 0.07 & 0.09 & 0.09 & 0.11 & 0.12 & 0.12 & 0.13 & 0.13 & 0.13 & 0.97 & 0.99 \\
        \bottomrule
    \end{tabularx}
    \vspace{0.5\baselineskip}
    \caption{\textbf{Normalized Effective Sample Size.} 
    We report the mean normalized ESS of \Ours{} along the denoising schedule and the resampling rate for each application. 
    }
    \label{tab:ess}
\end{table}

%% file: Tables/reward_sensitivity.tex
\begin{table}[h!]
    \centering
    \small
    \setlength{\tabcolsep}{3pt}
    \renewcommand{\arraystretch}{1.1}

    \begin{subtable}[b]{\linewidth}
        \centering
        \begin{tabular}{l l cccccc}
            \toprule
            \textbf{Task} & \textbf{Given Reward} & Mean & Std & Med. & 90th Pct. & Min & Max \\
            \midrule
            \textbf{Toxic Text Generation} & Toxicity Classifier & -9.465  & 1.470  & -10.041 & -8.046 & -10.364 & -0.020 \\
            \textbf{Regulatory DNA Design} & Enformer Activity   & 0.427   & 0.883  & 0.137   & 1.603  & -1.281  & 5.251 \\
            \textbf{Protein Designability} & Negative scRMSD     & -4.873  & 4.180  & -2.917  & -1.179 & -17.118 & -0.597 \\
            \textbf{dLLM Alignment} & Skywork Preference  & -17.815 & 10.254 & -19.000 & -3.778 & -43.000 & 17.625 \\
            \bottomrule
        \end{tabular}
        \caption{Base Model Reward Statistics}
        \label{tab:reward_stats}
    \end{subtable}

    \vspace{0.75\baselineskip}

    \begin{subtable}[b]{\linewidth}
        \centering
        \begin{tabularx}{0.75\linewidth}{l YYY YYY}
            \toprule
            & \multicolumn{3}{c}{\textbf{Given} $\uparrow$} & \multicolumn{3}{c}{\textbf{Heldout} $\uparrow$} \\
            \cmidrule(lr){2-4} \cmidrule(lr){5-7}
            \textbf{Given Reward} & $K=4$ & $K=8$ & $K=16$ & $K=4$ & $K=8$ & $K=16$ \\
            \midrule
            Original            & -5.483 & -4.057 & -3.036 & 0.387 & 0.460 & 0.547 \\
            Unbiased            & -4.203 & -2.781 & -1.843 & 0.793 & 0.900 & 0.970 \\
            Multilingual        & -1.847 & -0.867 & -0.588 & 0.633 & 0.710 & 0.800 \\
            Paradetox (Default) & -1.507 & -0.634 & -0.229 & 0.807 & 0.887 & 0.933 \\
            \midrule
            Paradetox + Noise   & -2.672 & -1.662 & -1.034 & 0.583 & 0.683 & 0.730 \\
            \bottomrule
        \end{tabularx}
        \caption{Toxic Text Generation with Different Reward Models}
        \label{tab:reward_models}
    \end{subtable}

    \caption{\textbf{Reward Sensitivity.} 
    (a) Reward statistics of the base model samples for each application. 
    (b) \Ours{} trained with different toxicity classifiers as the given reward, and with the default classifier perturbed by Gaussian noise. 
    }
    \label{tab:reward_sensitivity}
\end{table}

%% file: Tables/cost_analysis.tex
\begin{table}[h!]
    \centering
    \small
    \setlength{\tabcolsep}{2.9pt}
    \renewcommand{\arraystretch}{1.2}
    \begin{tabular}{l | cccccccc | c}
        \toprule
        & $T$ & \makecell{$c_\text{SMC}$ (s)} & \makecell{$c_\text{\Ours{}}$ (s)} & $c_\text{SMC}/c_\text{\Ours{}}$ & \makecell{$\Delta$ (s)} & $n_\text{pos}$ & $n_\text{neg}$ & \makecell{$C_\text{train}$ (h)} & $n^*$ \\
        \midrule
        \textbf{Toxic Text Generation} & 100 & 1.864   & 1.437 & $1.30\times$ & 0.427   & 5,120  & 20,480            & 6.73  & 56,727 \\
        \textbf{Regulatory DNA Design}        & 50  & 2.779   & 0.517 & $5.37\times$ & 2.262   & 1,024  & 4,096             & 1.08  & 1,716 \\
        \textbf{Protein Designability}    & 20  & 109.158 & 2.037 & $53.6\times$ & 107.121 & 192    & 1,920             & 6.00  & 202 \\
        \textbf{dLLM Alignment}       & 128 & 6.770   & 3.857 & $1.75\times$ & 2.912   & 25,600 & 12,800$^\dagger$  & 54.96 & 67,940 \\
        \bottomrule
    \end{tabular}
    \vspace{0.5\baselineskip}
    \caption{\textbf{Amortization Cost Analysis.} 
    We report the per-sample inference cost of SMC and \Ours{}, the upfront training cost of \Ours{}, and the break-even number of downstream samples $n^*$ beyond which \Ours{} becomes cheaper than SMC. 
    $^\dagger$For the negative sampling in dLLM alignment, we draw multiple samples from a single negative trajectory to save backbone forward passes. 
    }
    \label{tab:cost_analysis}
\end{table}

%% file: Figures/protein_main_qualitatives.tex
\begin{figure}[t]
\centering
\begin{tabularx}{\textwidth}{>{\centering\arraybackslash}p{2.8cm} |
                            >{\centering\arraybackslash}X}
\toprule
 & Samples \\
\midrule
Base~\cite{Wang:2025DPLM}       & \raisebox{-0.5\height}{\includegraphics[width=\linewidth]{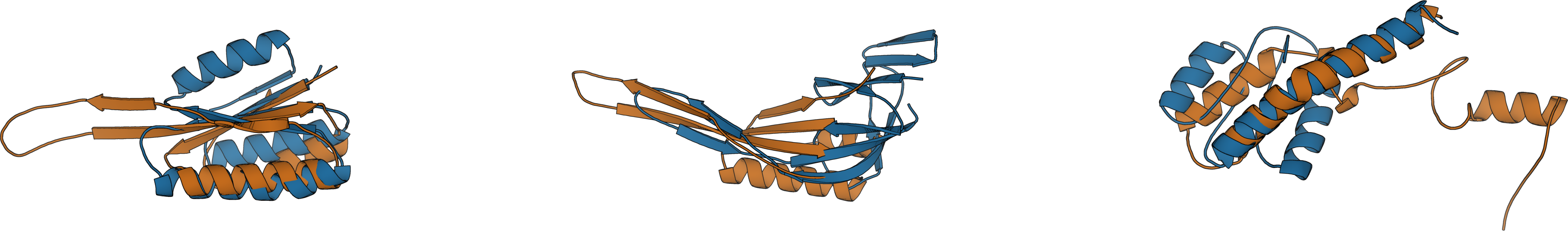}} \\
BoN        & \raisebox{-0.5\height}{\includegraphics[width=\linewidth]{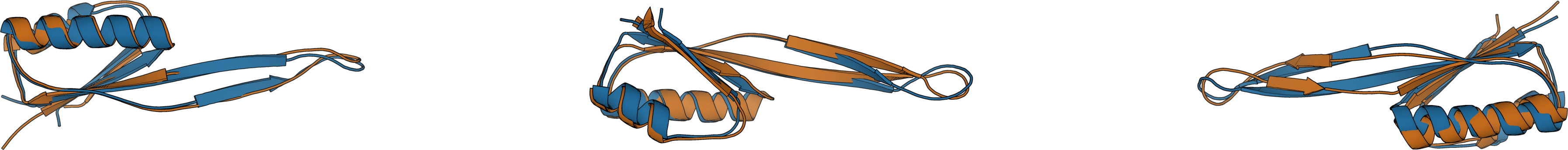}} \\
SMC ($M=1$)  & \raisebox{-0.5\height}{\includegraphics[width=\linewidth]{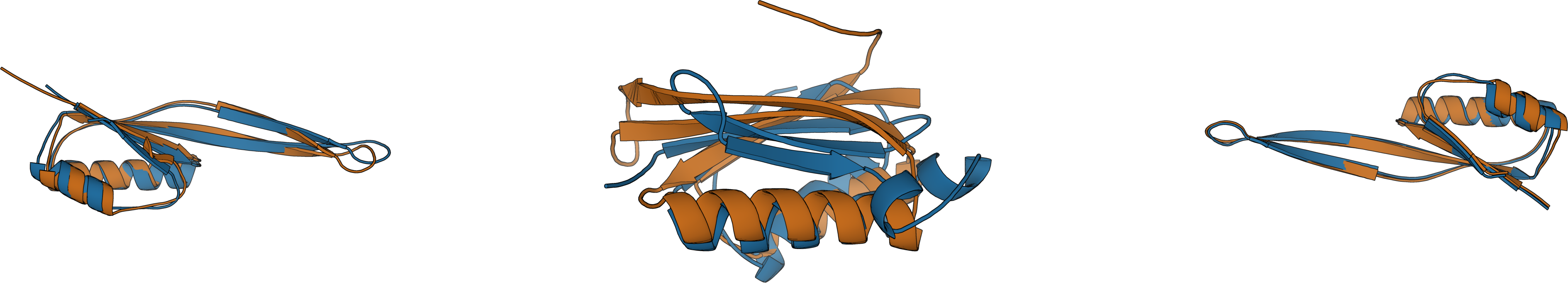}} \\
SMC ($M=4$)  & \raisebox{-0.5\height}{\includegraphics[width=\linewidth]{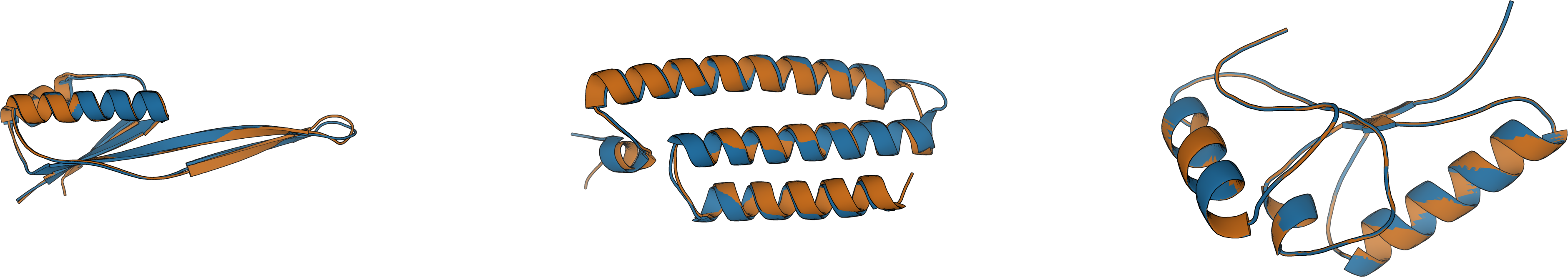}} \\
\svdd{}~\cite{Li2024:SVDD}       & \raisebox{-0.5\height}{\includegraphics[width=\linewidth]{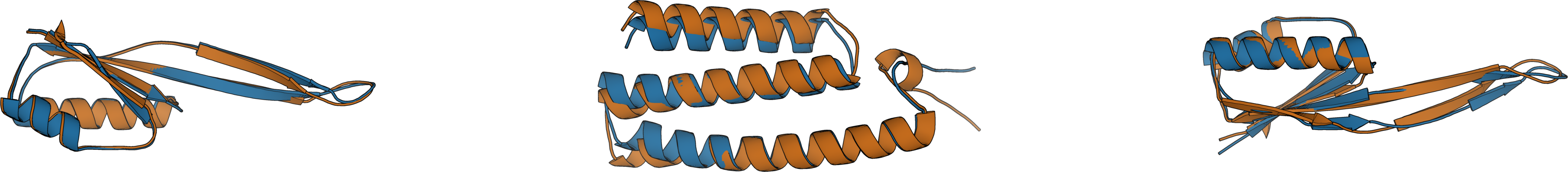}} \\
\Ours (Ours) & \raisebox{-0.5\height}{\includegraphics[width=\linewidth]{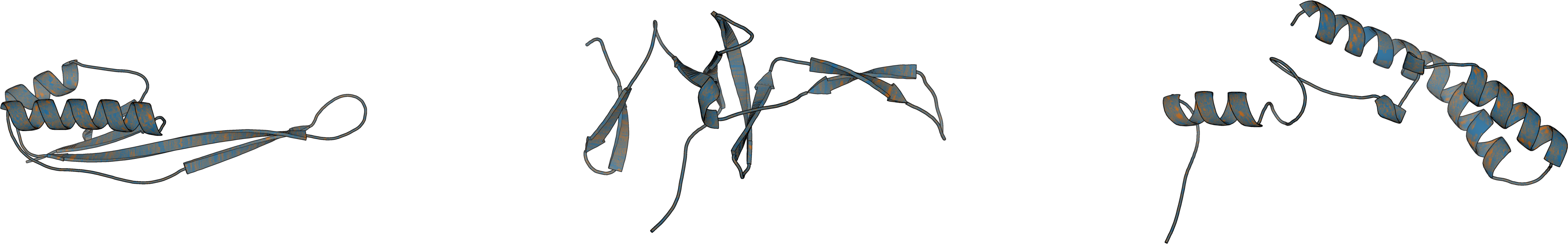}} \\
\bottomrule
\end{tabularx}
\caption{\textbf{Protein Qualitative Results.} We display the generated protein in blue and the refolded protein (using ESMFold) in orange. \Ours{} achieves designable proteins, as shown by the closely matching generated and refolded structures.}
\label{fig:protein-qualitatives}
\end{figure}

%% file: Tables/llada_alignment.tex
\begin{table}[h]
    \caption{\textbf{Diffusion Large Language Model Alignment.} We present qualitative examples generated using the LLaDA-8B-Instruct base model~\cite{nie:2025llada}. The results demonstrate that \Ours{} effectively generates high-quality responses that are well-aligned with user preferences.}
    \vspace{0.6em}
    \label{tab:llada_qual_start}
    \centering
    \small
    \begin{tabularx}{\linewidth}{LLL}
        \toprule
        \rowcolor{lightgray!50}
        Prompt & Base~\cite{nie:2025llada} & BoN \\
        \hline
        \noalign{\vspace{0.5em}}
        Write a friendly e-mail saying, that I won't be able to join todays meeting. Make up a very understandable reason, that's serious enough but won't lead to awkward questions from my coworkers tomorrow.
        &
        Hello Friend,\newline
        I hope this email finds you well. I accelerating we have a meeting scheduled for today, and somewhat a brown nosing family member, so informed me that you won't be able to attend the meeting.\par\medskip
        According to them, you wouldn't quite find out until the morning effects passed bad and therefore we needed to attend to an emergency, However this may be time to write, but I'm thoughtful apologies for your absence and won't report\begin{CJK}{UTF8}{gbsn}突发事件\end{CJK}, do anything else, as you'll need to make it up at your earliest convenience.
        &
        Subject: Today's Meeting Attendance\par\medskip
        Hi Team,\newline
        I hope this email finds you well. I need to inform you regarding attendance for meeting. Unfortunately, I have last minute family call, which doesn't exactly fit me to be feeling. It seems a few hours away, a family family situation really needs a case. If the time and any issues seems joining today, we can still discuss and review everything later?\par\medskip
        Thanks for your flexibility and I hope this has been misunderstood.\newline
        Best regards,\newline
        [Your Name] \\
        \midrule
        \rowcolor{lightgray!50}
        SMC & \svdd{}~\cite{Li2024:SVDD} & \Oursbf{} \\
        \hline
        \noalign{\vspace{0.5em}}
        Hello [ coworker],\par\medskip
        I hope this not napping you but I will have to be unable to join today's meeting. Recently I have been having a difficult time with coughing due to a known chronic rh life\begin{CJK}{UTF8}{gbsn}说她\end{CJK} that I have had been struggling for for 10/15 years now strums and job\begin{CJK}{UTF8}{gbsn}见他\end{CJK}didn't you have a professional where vet advised me against unnecessary meetings as soon as possible.\par\medskip
        I understand for what is very necessary for business I have deferred my meeting and will have all related documents taking care of for discuss asap.\par\medskip
        Kindest of regards,\newline
        [Your Name]
        &
        Subject: Not Analytics Today\par\medskip
        Dear [Colleagues's Names],\par\medskip
        I hope this message is finding you well. Unfortunately, I will not be able to take part in today's meeting due to some health issues. I am missing critical sales data for\begin{CJK}{UTF8}{gbsn}先股权投资\end{CJK} and will complete and submit it next week. I will be able to present internal\_with\_ids data and will provide the sales update tomorrow - until\begin{CJK}{UTF8}{gbsn}先股权投资\end{CJK}'s data arrives before decisions are implemented looking forward.\par\medskip
        Let's make it simple and wait for the sales update tomorrow. Thank you for your understanding.\par\medskip
        Best regards,\newline
        [Your Name]
        &
        Hi [Coworker's Name],\par\medskip
        I hope this email finds you well. I wanted to kindly inform you that I won't be able to join today's meeting. Unfortunately, earlier this afternoon, I came across some important personal matters that need to be attended to immediately. I hope to have them addressed as soon as possible.\par\medskip
        If you have any specific details or concerns regarding today's meeting, please let me know. I'll update you all once I am available.\par\medskip
        Thank you for your understanding, and I look forward to our ongoing discussions.\par\medskip
        Best regards,\newline
        [Your Full Name] \\
        \bottomrule
    \end{tabularx}
\end{table}
\clearpage

\begin{table}[h]
    \caption{\textbf{Diffusion Large Language Model Alignment.} We present qualitative examples generated using the LLaDA-8B-Instruct base model~\cite{nie:2025llada}. The results demonstrate that \Ours{} effectively generates high-quality responses that are well-aligned with user preferences.}
    \vspace{0.6em}
    \centering
    \small
    \begin{tabularx}{\linewidth}{LLL}
        \toprule
        \rowcolor{lightgray!50}
        Prompt & Base~\cite{nie:2025llada} & BoN \\
        \hline
        \noalign{\vspace{0.5em}}
        Can you analyze this part of a numerical sequence and provide me the next 3 numbers in it?\newline
        Sequence: 2, 4, 8, 16, 32
        &
        Certainly! Observing the sequence 2, 4, 8, 16, 32, it appears that each number is obtained by multiplying the previous number by 2. Let's check this pattern:\par\medskip
        1. \(4 = 2 \times 2\)\newline
        2. However \(8 = 4 \times 2\) and \(16 = 8 \times 2\), indicating we can see a clear progression by the power of 2\par\medskip
        4. The next number\newline
        The next progression would be 2, 4, 8, and 16.
        &
        Sure! The sequence you provided is a geometric progression where each number is obtained by multiplying the previous number by 2. In other words, the factor that contributes to this multiplication is \(2\).\par\medskip
        So, let's calculate the next three numbers accordingly:\par\medskip
        1. \(16 \times 2 = 32\)\newline
        2. \(32 \times 2 = 64\)\newline
        3. \(64 \times 2 = 128\)\par\medskip
        Thus, the next three numbers in the sequence are 32, 64, and 128. \\
        \midrule
        \rowcolor{lightgray!50}
        SMC & \svdd{}~\cite{Li2024:SVDD} & \Oursbf{} \\
        \hline
        \noalign{\vspace{0.5em}}
        Based on the sequence provided, it appears that each subsequent number is obtained by multiplying the previous number by 2. If we are to continue this pattern, let's double the last number of the sequence to find the next 3 numbers in sequence:\par\medskip
        1. 32 (from sequence)\newline
        2. 32 x 2 = 64\newline
        3. 64 x 2 = 128\par\medskip
        So, the next 3 numbers in the sequence would be:\par\medskip
        - 32\newline
        - 64\newline
        - 128
        &
        Yes, I'd be happy to analyze this numerical sequence.\par\medskip
        Let's look at the pattern in the sequence:\par\medskip
        The sequence given is: 2, 4, 8, 16, 32.\par\medskip
        The pattern here is that each term is two times the previous term. Therefore, the next three numbers in the sequence would be:\par\medskip
        1. \( 32 * 2 = 64 \)\newline
        2. \( 64 * 2 = 128 \)\newline
        3. \( 64 * 2 = 256 \)
        &
        The sequence you've provided is a geometric sequence with a common ratio of 2. The next three numbers in the sequence would be:\par\medskip
        64, 128, 256 \\
        \bottomrule
    \end{tabularx}
\end{table}
\clearpage

\begin{table}[h]
    \caption{\textbf{Diffusion Large Language Model Alignment.} We present qualitative examples generated using the LLaDA-8B-Instruct base model~\cite{nie:2025llada}. The results demonstrate that \Ours{} effectively generates high-quality responses that are well-aligned with user preferences.}
    \vspace{0.6em}
    \label{tab:llada_qual_joke}
    \centering
    \small
    \begin{tabularx}{\linewidth}{LLL}
        \toprule
        \rowcolor{lightgray!50}
        Prompt & Base~\cite{nie:2025llada} & BoN \\
        \hline
        \noalign{\vspace{0.5em}}
        Tell me a joke that involves a word-play with the word "rain"
        &
        Here's a classic joke that involves a word-play with the word "rain":\par\medskip
  
        Why did the umbrella quit over the chicken?\par\medskip
          
        Because it let it rain!
        &
        Why did the landlord try to get people to pay the RENT  on time - it was not the kind he wears. \\
        \midrule
        \rowcolor{lightgray!50}
        SMC & \svdd{}~\cite{Li2024:SVDD} & \Oursbf{} \\
        \hline
        \noalign{\vspace{0.5em}}
        Here's a word-play using the word "rain":\par\medskip
  
        Why did the rain go to "invite"? Because it was a little wet.\par\medskip
          
        This joke uses play on the words, the quiet about the rain can make it cramp or wet, and the idea of "inviting" the situation to the rain where it's making things wet or slippery, just like one gets invited to having to be in a wet location. The word "INVite" happens to be a title, which adds to the topical humor.
        &
        Why did the umbrella quit its job? \par\medskip
  
        Because it needed some pick-me-up!
        &
        Why did the rain go to the doctor?\par\medskip
  
        Because it was feeling drained!\\
        \bottomrule
    \end{tabularx}
\end{table}
\clearpage

\begin{table}[h]
    \caption{\textbf{Diffusion Large Language Model Alignment.} We present qualitative examples generated using the LLaDA-8B-Instruct base model~\cite{nie:2025llada}. The results demonstrate that \Ours{} effectively generates high-quality responses that are well-aligned with user preferences.}
    \vspace{0.6em}
    \label{tab:llada_qual_transcription}
    \centering
    \small
    \begin{tabularx}{\linewidth}{LLL}
        \toprule
        \rowcolor{lightgray!50}
        Prompt & Base~\cite{nie:2025llada} & BoN \\
        \hline
        \noalign{\vspace{0.5em}}
        Correct the transcription of an excerpt containing errors.\par\medskip
        I got got charged interest on ly credit card but I paid my pull balance one day due date. I not missed a pavement year yet. Man you reverse the interest charge?
        &
        Errors I'm using \$16. 10 but I paid off my balance one day before the due date. I haven't missed a payment yet. Can you refund this interest charge?

        &
        I was charged interest on a past due credit card balance, but I paid my balance one day before the due date. I have not made any late payments yet. Can you please waive the interest charge?\\
        \midrule
        \rowcolor{lightgray!50}
        SMC & \svdd{}~\cite{Li2024:SVDD} & \Oursbf{} \\
        \hline
        \noalign{\vspace{0.5em}}
        I was charged interest on my credit card, but I paid my full balance on the day of the due date. I haven't missed a single payment yet. Can you reverse the interest charge?
        &
        Here's the corrected transcription:\par\medskip
  
        I got charged interest on my credit card, but I paid the balance at the due date. I didn't miss a payment yet. Can you reverse the interest charge?\par\bigskip

        Errors changed:\newline
        - "ly" was removed\newline
        - "accense" was used as extra emphasizes on correct spelling\newline
        - "pull" was corrected to "balance"\newline
        - "not" was corrected to "didn't"\newline
        - "out of punctuation" was swapped\newline
        - Removed split sentences and merged poistuations
        &
        I was charged interest on my credit card, but I paid my full balance on the due date. I haven't missed a payment this year. Can you reverse the interest charge?\\
        \bottomrule
    \end{tabularx}
\end{table}
\clearpage

\begin{table}[h]
    \caption{\textbf{Diffusion Large Language Model Alignment.} We present qualitative examples generated using the LLaDA-8B-Instruct base model~\cite{nie:2025llada}. The results demonstrate that \Ours{} effectively generates high-quality responses that are well-aligned with user preferences.}
    \vspace{0.6em}
    \label{tab:llada_qual_sqrt}
    \centering
    \small
    \begin{tabularx}{\linewidth}{LLL}
        \toprule
        \rowcolor{lightgray!50}
        Prompt & Base~\cite{nie:2025llada} & BoN \\
        \hline
        \noalign{\vspace{0.5em}}
        If $\sqrt{3x-5}=2$, find all possible values of $x$.
        &
        Squaring both sides of the given equation, we get\newline
        \[ 3x-5 = 4. \] \newline
        Subtracting $5$ from both sides yields\newline
        \[ 3x = \boxed{4 +5} = \boxed{9}. \]\newline
        Then to satisfy $x$, $3x\geq{5}$, making $x$ satisfy the original equation is \newline
        \[ x = \boxed{\frac{9}{3}}. \]\newline
        Therefore, there is only one possible value of \( x \), that value is \(x = \boxed{\frac{9}{3}}\).
        &
        We have the equation $\sqrt{3x-5} = 2$. \newline
        Squaring both sides, we get $(\sqrt{3x-5})^2 = 2^2$.\newline
        Simplifying, we have $3x-5 = 4$.\newline
        Subtracting 5 from both sides gives us $3x = 13$.\newline
        Dividing both sides by 3 gives us $x = \frac{13}{3}$.\newline
        So the possible value of $x$ is $\boxed{\frac{3}{13}}$.The answer is: $\frac{13}{3}$\\
        \midrule
        \rowcolor{lightgray!50}
        SMC & \svdd{}~\cite{Li2024:SVDD} & \Oursbf{} \\
        \hline
        \noalign{\vspace{0.5em}}
        We have the equation $\sqrt{3x - 5} = 2$. Squaring both sides, we get $3x - 5 = 9$. Solving for $x$, we get the unique value for $x = \boxed{\frac{14}{3}}$.
        &
        Given $\sqrt{3x-5}=2$, squaring both sides, we get \[\sqrt{3x-5}=(2^2) \implies 4.
        \]\newline
        Thus, we have \[\newline
        3x-5=4 \Rightarrow 3x=9.\newline
        \]\newline
        Dividing both sides by 3, we get \[\newline
        x=\boxed{\frac{29}{3}}.\newline
        \]
        &
        Given the equation \(\sqrt{3x-5} = 2\).\par\medskip
  
        To eliminate the square root, we will square both sides of the equation.\par\medskip
          
        \[\newline
        3x - 5 = 4\newline
        \]\par\medskip
          
        Next, we solve for \(x\). Add 5 to both sides:\par\medskip
          
        \[\newline
        3x = 9\newline
        \]\par\medskip
          
        Then, divide both sides by 3:\par\medskip
          
        \[\newline
        x = 3\newline
        \]\par\medskip
          
        Thus, the only possible value of \(x\) is:\par\medskip
          
        \[\newline
        \boxed{3}\newline
        \]
        \\
        \bottomrule
    \end{tabularx}
\end{table}
\clearpage

\begin{table}[h]
    \caption{\textbf{Diffusion Large Language Model Alignment.} We present qualitative examples generated using the LLaDA-8B-Instruct base model~\cite{nie:2025llada}. The results demonstrate that \Ours{} effectively generates high-quality responses that are well-aligned with user preferences.}
    \vspace{0.6em}
    \label{tab:llada_qual_inequality}
    \centering
    \small
    \begin{tabularx}{\linewidth}{LLL}
        \toprule
        \rowcolor{lightgray!50}
        Prompt & Base~\cite{nie:2025llada} & BoN \\
        \hline
        \noalign{\vspace{0.5em}}
        For what values of $x$ is it true that $x^2 - 5x - 4 \le 10$? Express your answer in interval notation.
        &
        This inequality is $x^2 \le 14$.\newline
        We can factor our inequality as follows,\newline
        \[(x-2)(x-7)\ne0.\]\newline
        We have ignored the roots for the quadratic like we did earlier.\newline
        So, the inequality holds only if $x\ge7$.\newline
        Since the quadratic is symmetric around $14$ the same for the roots $x\ge10$ and boundary $x=-2$ for which $x$ is negative, $x\ge7$ is the only solution.\newline
        So, the answer is $\boxed{\{\infty,7\}}$.
        &
        After subtracting 10 from each side, we have $x^2 - 5x - 4 - 10 \le 0$, which we can rewrite as: $x^2 - x - 14 \le 0$. This is a quadratic inequality, and can easily be solved by finding factors:  $$(x - 4)(x + 3) \le 0.$$ The roots are $x=-4$ and $x=3$, so we box the final interval:\par\medskip
  
        \boxed{ (-\infty, -4]\cup [3, \infty)}. .\\
        \midrule
        \rowcolor{lightgray!50}
        SMC & \svdd{}~\cite{Li2024:SVDD} & \Oursbf{} \\
        \hline
        \noalign{\vspace{0.5em}}
        The given inequality becomes $$ x^2 - 5x - 4 - 10 \le 0,$$ or $$
        x^2 - 5x - 14 \le 0.$$\newline
        We know that $x=14$ or $x=-7$, so $(x+1)(x-7)$ is a factor.\newline
        We quotient to find\begin{CJK}{UTF8}{gbsn}不同意\end{CJK} infants $(x+2)(4x+7)$. Thus, our answer is $$x \in \boxed{-\frac{7}{2} \cup (-7, -2) \cup (-2, 7]}.$$
        &
        Subtracting $10$ from both sides gives \par\medskip
  
        \[\newline
        x^2 - 5x - 4 \le 0. \newline
        \] \par\medskip
          
        Factoring this inequality gives \par\medskip
          
        \[\newline
        (x + 2)(x - 7) \le 0. \newline
        \]\par\medskip
          
        Therefore, this is true when $x$ is to the left of $x = -2$ or right of $x = 7.$ In interval notation, this is \par\medskip
          
        \[\newline
        \boxed{(-\infty, -2] \cup [7, \infty)}.\newline
        \]
        &
        We rewrite the inequality $x^2 - 5x - 4 - 10\le 0$ as $x^2 - 5x - 14 \le 0.$ This is equivalent to $(x - 7)(x + 2) \le 0,$ which holds for $x \in \boxed{[-2, 7]}.$
        \\
        \bottomrule
    \end{tabularx}
\end{table}
\clearpage

\begin{table}[h]
    \caption{\textbf{Diffusion Large Language Model Alignment.} We present qualitative examples generated using the LLaDA-8B-Instruct base model~\cite{nie:2025llada}. The results demonstrate that \Ours{} effectively generates high-quality responses that are well-aligned with user preferences.}
    \vspace{0.6em}
    \label{tab:llada_qual_last}
    \centering
    \small
    \begin{tabularx}{\linewidth}{LLL}
        \toprule
        \rowcolor{lightgray!50}
        Prompt & Base~\cite{nie:2025llada} & BoN \\
        \hline
        \noalign{\vspace{0.5em}}
        Write a Python function `rescale\_to\_unit(numbers: List[float]) -> List[float]` to solve the following problem:\newline
        Given list of numbers (of at least two elements), apply a linear transform to that list,\newline
        such that the smallest number will become 0 and the largest will become 1\newline
        >>> rescale\_to\_unit([1.0, 2.0, 3.0, 4.0, 5.0])\newline
        [0.0, 0.25, 0.5, 0.75, 1.0]
        &
        Here are some common functions whose data types areHP or variable-precision:\newline
        >>> from typing import Union, max\newline
        >>> where - Union[int, float, Optional[float]] type of t.item (\newline
        in case\newline
        \quad of t is int or\newline
        \quad is float\par\medskip
          
        This reveals that our answer makes sense, since:\newline
        >>> rescale\_to\_unit([1.0, 2.0, 3.0, 4.0, 5.0]) -> 0.0, 0.25, 0.5, 0.75, 1.0 (or
        &
        Here is a code that does as follows:\par\bigskip

        ```\newline
            \quad from math import\begin{CJK}{UTF8}{gbsn}入门\end{CJK}o.  \# Use the math module\par\medskip
          
            \quad a\_min = min(numbers)\par\medskip
          
            \quad a\_max = max(numbers)\par\medskip
          
            \quad if a\_max < a\_min:\par\medskip
          
                \quad\quad raise ValueError("The input list must contain non empty elements")  \# Avoiding two zeroes\par\medskip
          
            \quad \# res: List of ListMember(Any, "0")\par\medskip
          
            \quad res = [(num - a\_min) / (a\_max - a\_min) for num in numbers]\par\medskip
          
            \quad return res\par\medskip
          
        ```\\
        \midrule
        \rowcolor{lightgray!50}
        SMC & \svdd{}~\cite{Li2024:SVDD} & \Oursbf{} \\
        \hline
        \noalign{\vspace{0.5em}}
        Here is an Answer using list comprehension (`Python 3.5`)\newline
        from List import List, Tuple\par\medskip
          
        def rescale\_to\_unit(numbers: List[float]) -> List[float]:\newline
            \quad smallest, largest = min(numbers), max(numbers)\newline
            \quad return list((number, number - smallest) / largest for number in Numbers)\newline
        wavex[int]\newline
        The `vel`, numbers` computes the minimum and maximum in the list.  The next line uses these endpoints to\begin{CJK}{UTF8}{gbsn}如此在总的话就\end{CJK}haped return then returns the minized version of the numbers before they were transformed (min, naturally)
        &
        def rescale\_to\_unit(numbers):\newline
        ```python\newline
        (numbers: []) // numbers -> float:\newline
            \quad if len(numbers) < 2:\newline
                \quad \quad raise ValueError("List must contain at least two elements")\par\medskip
              
            \quad return [(val - min\_val) / max\_val for val in numbers]\newline
        ```\newline
        >>> rescale\_to\_unit([1.0, 2.0, 3.0, 4.0, 5.0])\newline
        [0.0, 0.0, 0.25, 0.5, 0.75]
        &
        from typing import List\par\bigskip

        def rescale\_to\_unit(numbers: List[float]) -> List[float]:\newline
           \quad min\_num = min(numbers)\newline
           \quad max\_num = max(numbers)\newline
           \quad return [(x - min\_num) / (max\_num - min\_num) for x in numbers]
        \\
        \bottomrule
    \end{tabularx}
\end{table}
\clearpage

%% file: checklist.tex
\section*{NeurIPS Paper Checklist}

\begin{enumerate}

\item {\bf Claims}
    \item[] Question: Do the main claims made in the abstract and introduction accurately reflect the paper's contributions and scope?
    \item[] Answer: \answerYes{} %
    \item[] Justification: We propose contrastive twist learning objective in discrete diffusion which outperforms the baselines across diverse applications. 

\item {\bf Limitations}
    \item[] Question: Does the paper discuss the limitations of the work performed by the authors?
    \item[] Answer: \answerYes{} %
    \item[] Justification: We discuss limitations and future works in the paper. 

\item {\bf Theory assumptions and proofs}
    \item[] Question: For each theoretical result, does the paper provide the full set of assumptions and a complete (and correct) proof?
    \item[] Answer: \answerYes{} %
    \item[] Justification: We present all proofs and derivations in Appendix.~\ref{sec:proof}. 

    \item {\bf Experimental result reproducibility}
    \item[] Question: Does the paper fully disclose all the information needed to reproduce the main experimental results of the paper to the extent that it affects the main claims and/or conclusions of the paper (regardless of whether the code and data are provided or not)?
    \item[] Answer: \answerYes{} %
    \item[] Justification: We present algorithms, hyperparameters, and implementation details. The pretrained models are publicly available. 

\item {\bf Open access to data and code}
    \item[] Question: Does the paper provide open access to the data and code, with sufficient instructions to faithfully reproduce the main experimental results, as described in supplemental material?
    \item[] Answer: \answerNo{} %
    \item[] Justification: We will release the code publicly upon acceptance. 

\item {\bf Experimental setting/details}
    \item[] Question: Does the paper specify all the training and test details (e.g., data splits, hyperparameters, how they were chosen, type of optimizer) necessary to understand the results?
    \item[] Answer: \answerYes{} %
    \item[] Justification: We present experiment setups and implementation details of all applications. 

\item {\bf Experiment statistical significance}
    \item[] Question: Does the paper report error bars suitably and correctly defined or other appropriate information about the statistical significance of the experiments?
    \item[] Answer: \answerNo{} %
    \item[] Justification:
    We report the mean metrics averaged across multiple independent runs. 
    Large-scale experiments were infeasible due to computational constraints. 

\item {\bf Experiments compute resources}
    \item[] Question: For each experiment, does the paper provide sufficient information on the computer resources (type of compute workers, memory, time of execution) needed to reproduce the experiments?
    \item[] Answer: \answerYes{} %
    \item[] Justification: We discuss computational overhead and training time in Appendix. 
    
\item {\bf Code of ethics}
    \item[] Question: Does the research conducted in the paper conform, in every respect, with the NeurIPS Code of Ethics \url{https://neurips.cc/public/EthicsGuidelines}?
    \item[] Answer: \answerYes{} %
    \item[] Justification: We conform to the NeurIPS Code of Ethics. 

\item {\bf Broader impacts}
    \item[] Question: Does the paper discuss both potential positive societal impacts and negative societal impacts of the work performed?
    \item[] Answer: \answerYes{} %
    \item[] Justification: We discuss societal impact of our applications. 
    
\item {\bf Safeguards}
    \item[] Question: Does the paper describe safeguards that have been put in place for responsible release of data or models that have a high risk for misuse (e.g., pre-trained language models, image generators, or scraped datasets)?
    \item[] Answer: \answerNA{} %
    \item[] Justification: No high risk data or model is used. 

\item {\bf Licenses for existing assets}
    \item[] Question: Are the creators or original owners of assets (e.g., code, data, models), used in the paper, properly credited and are the license and terms of use explicitly mentioned and properly respected?
    \item[] Answer: \answerYes{} %
    \item[] Justification: We include all citations and references. 

\item {\bf New assets}
    \item[] Question: Are new assets introduced in the paper well documented and is the documentation provided alongside the assets?
    \item[] Answer: \answerNA{} %
    \item[] Justification: We do not present new assets. 

\item {\bf Crowdsourcing and research with human subjects}
    \item[] Question: For crowdsourcing experiments and research with human subjects, does the paper include the full text of instructions given to participants and screenshots, if applicable, as well as details about compensation (if any)? 
    \item[] Answer: \answerNA{} %
    \item[] Justification: We do not include human subject studies or crowdsourcing. 

\item {\bf Institutional review board (IRB) approvals or equivalent for research with human subjects}
    \item[] Question: Does the paper describe potential risks incurred by study participants, whether such risks were disclosed to the subjects, and whether Institutional Review Board (IRB) approvals (or an equivalent approval/review based on the requirements of your country or institution) were obtained?
    \item[] Answer: \answerNA{} %
    \item[] Justification: We do not experiments on human subjects. 

\item {\bf Declaration of LLM usage}
    \item[] Question: Does the paper describe the usage of LLMs if it is an important, original, or non-standard component of the core methods in this research? Note that if the LLM is used only for writing, editing, or formatting purposes and does \emph{not} impact the core methodology, scientific rigor, or originality of the research, declaration is not required.
    \item[] Answer: \answerNA{} %
    \item[] Justification: LLM is not used in proposition of the core method. 

\end{enumerate}